\documentclass[journal, compsoc]{IEEEtran}

\usepackage{cite}
\usepackage{amsmath,amssymb,amsfonts}
\usepackage{algorithmic}
\usepackage{graphicx}
\usepackage{textcomp}
\usepackage{url}
\usepackage{xcolor}
\usepackage{times}
\usepackage{epsfig}
\usepackage{amssymb}
\usepackage{fixltx2e}
\usepackage{listings}
\usepackage{booktabs} 
\usepackage{multirow}
\usepackage{gensymb}
\usepackage{mathtools}
\usepackage{adjustbox}
\usepackage{framed}
\usepackage{latexsym}
\usepackage{amsfonts}
\usepackage{dsfont}
\usepackage{float}

\usepackage{xcolor}
\definecolor{DarkGreen}{rgb}{0.2,0.5,0.2} 

\makeatletter
\AtBeginDocument{\def\@citecolor{DarkGreen}}
\makeatother

\usepackage{arydshln}
\newcommand{\etal}{\textit{et al}.}

\usepackage[pagebackref=true,breaklinks=true,letterpaper=true,colorlinks,bookmarks=false, urlcolor=DarkGreen, citecolor=DarkGreen]{hyperref}

\begin{document}

\title{GANalyzer: Analysis and Manipulation of GANs Latent Space for Controllable Face Synthesis}

\author{Ali~Pourramezan~Fard, 
        Mohammad H. Mahoor, \IEEEmembership{Senior Member, IEEE},
        Sarah Ariel Lamer, and
        Timothy Sweeny

\IEEEcompsocitemizethanks{
\IEEEcompsocthanksitem Ali~Pourramezan~Fard and Mohammad H. Mahoor are with the Ritchie School of Engineering and Computer Science, University of Denver, Denver,
CO, 80208 .\protect\\
E-mail: Ali.Pourramezanfard@du.edu, Mohammad.Mahoor@du.edu

\IEEEcompsocthanksitem  Sarah Ariel Lamer is with the Department of Psychology, University of Tennessee, Knoxville, TN, 37996 .\protect\\
E-mail: slamer@utk.edu

\IEEEcompsocthanksitem Timothy Sweeny is with the College of Arts, Humanities and Social Sciences, University of Denver, Denver,
CO, 80208 .\protect\\
E-mail: timothy.sweeny@du.edu

}


}


\maketitle

\begin{abstract}
Generative Adversarial Networks (GANs) are capable of synthesizing high-quality facial images. Despite their success, GANs do not provide any information about the relationship between the input vectors and the generated images. Currently, facial GANs are trained on imbalanced datasets, which generate less diverse images. For example, more than 77\% of 100K images that we randomly synthesized using the StyleGAN3 are classified as Happy, and only around 3\% are Angry. The problem even becomes worse when a mixture of facial attributes is desired: less than 1\% of the generated samples are Angry Woman, and only around 2\% are Happy Black. To address these problems, this paper proposes a framework, called GANalyzer, for the analysis, and manipulation of the latent space of well-trained GANs. GANalyzer consists of a set of transformation functions designed to manipulate latent vectors for a specific facial attribute such as facial Expression, Age, Gender, and Race. We analyze facial attribute entanglement in the latent space of GANs and apply the proposed transformation for editing the disentangled facial attributes. Our experimental results demonstrate the strength of GANalyzer in editing facial attributes and generating any desired faces. We also create and release a balanced photo-realistic human face dataset. Our code is publicly available \href{https://github.com/aliprf/GANalyzer/}{here}.

\end{abstract}

\begin{IEEEkeywords}
Generative adversarial network, face editing, latent space interpretation, facial attribute editing, transformation\end{IEEEkeywords}

\IEEEpeerreviewmaketitle

\section{Introduction}
\IEEEPARstart{R}{ecently}, we have witnessed great success and advancement in the quality of the images being synthesized by Generative Adversarial Networks (GANs)~\cite{goodfellow2020generative}. GANs learn a mapping between a random distribution and a distribution of real data, using adversarial training. As a result, GANs can generate photo-realistic images from randomly sampled vectors from latent space.

\begin{figure}[t]
  \centering
  \includegraphics[width=\columnwidth]{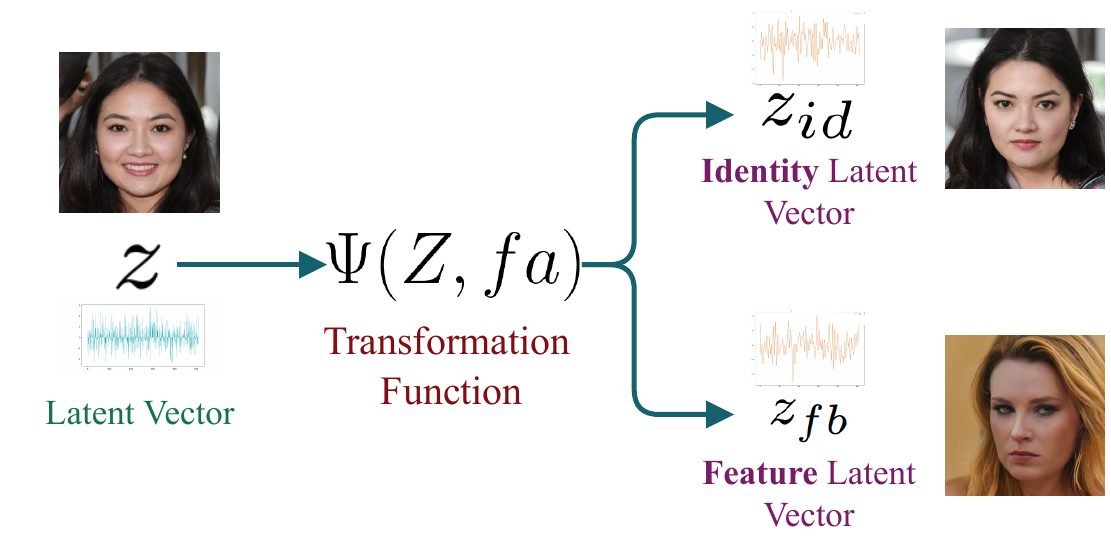}
  \caption{Our proposed transformation function $\Psi$ gets an arbitrary latent vector $z$, a facial attribute $fa$, and manipulates $z$ to generate the latent vectors $z_{fb}$, and $z_{id}$. While the latent vector $z_{id}$ is used for facial attribute editing, the feature latent vector $z_{fb}$ is used for feature-based synthesis. }
  \label{fig:transformation_function}
\end{figure}
%

Despite the ability of GANs in the synthesis of high-fidelity images, GANs can not provide any information about the relation between the facial attributes and features of the synthesized images, and each element of the corresponding latent vector~\cite{shen2020interfacegan}. Hence, we are not able to utilize such coding information to control the facial attributes and features of the generated image. Interpreting the latent space of GANs would provide us with control over the attributes of the generated images. For human face synthesis, an ideal interpretation of the latent space should provide tools for both \textit{Facial Attribute Editing} and \textit{Feature-Based Synthesis}.

Facial attribute editing~\cite{yangl2m} is an interesting research topic with a wide range of real-world applications such as entertainment, auxiliary psychiatric treatment, and data augmentation. The main goal of facial attribute editing is to preserve a person's identity while changing a set of specific attributes of their face. Contrary to facial attribute editing, we introduce a concept and call it \textit{feature-based synthesis}, where the goal is to synthesize photo-realistic human faces with specific facial attributes. More specifically, instead of modifying the facial attributes of a \textit{previously} generated image, in feature-based synthesis, we want to generate human faces that have specific facial attributes (\textit{e.g.} perceived facial expression, age, gender, race). As an example, in facial attribute editing, firstly we generate an image and then edit its facial expression and gender, while in feature-based synthesis, we can synthesize images with a specific facial expression, and gender.

One of the main applications of such an approach is creating diverse, and balanced datasets that can be used in other domains, such as facial expression recognition, age estimation, ethnicity recognition, and a wide variety of medical and psychological research.

\begin{figure}[t]
  \centering
  \includegraphics[width=\columnwidth]{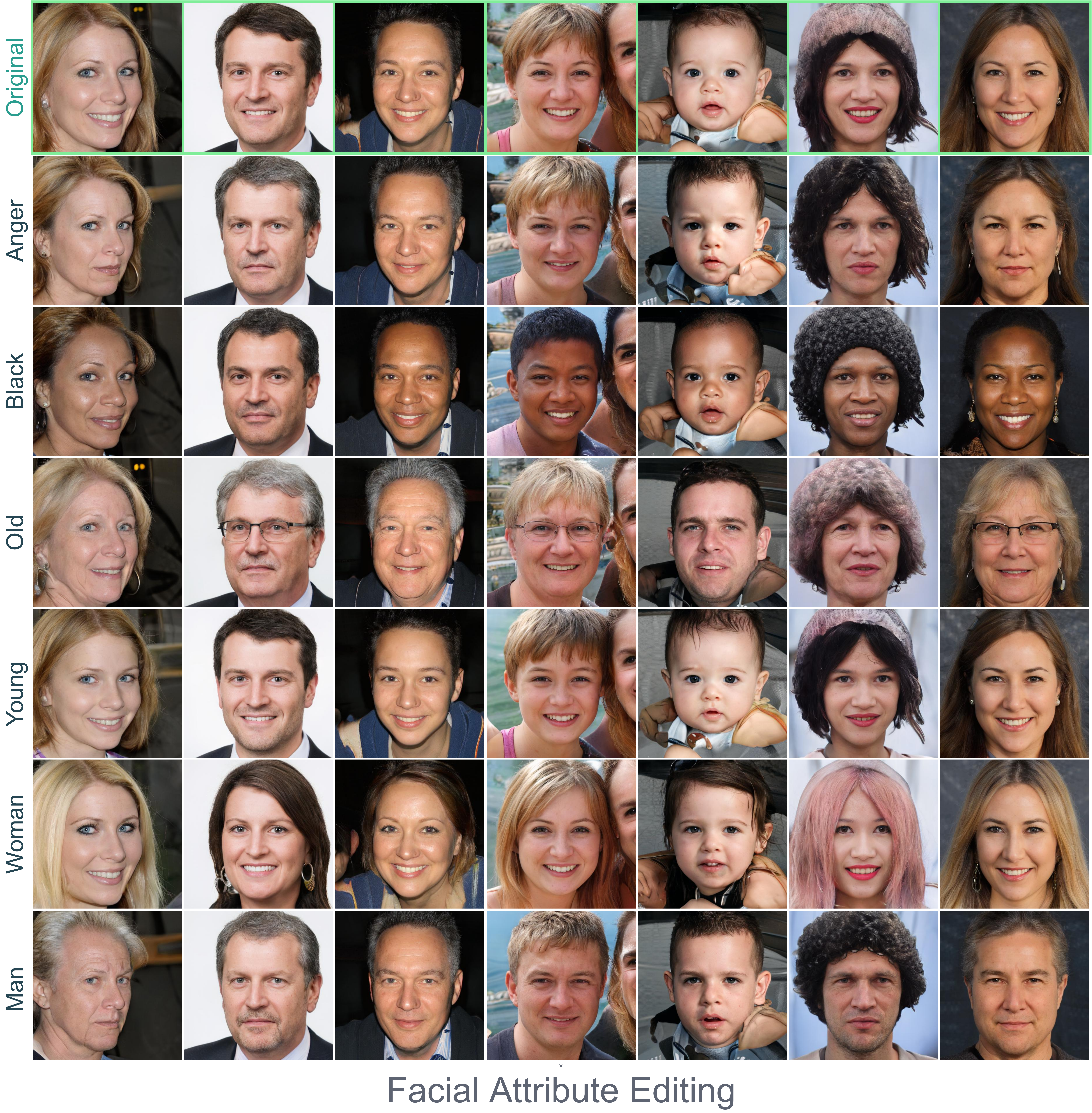}
  \caption{ The figure shows a performance of our proposed transformation function $\Psi$ for facial attribute editing. }
  \label{fig:single_all_attribute}
\end{figure}
%

In this paper, we propose a framework called GANalyzer to interpret and analyze the latent space of GANs for \textit{both} facial attribute editing and feature-based image synthesis. GANalyzer is designed to analyze the latent space of well-trained GANs, and hence, learn how the manipulation of a latent vector could affect the attributes of the generated images. As Fig.~\ref{fig:single_all_attribute} shows, from the facial attribute editing perspective, GANalyzer is capable of modifying a latent vector such that only a specific attribute of the corresponding image is changed (\textit{e.g.} modifying a person perceived as a woman to instead be perceived as a man, while preserving the other facial attributes such as facial expression, age, \textit{etc.}). Likewise, from feature-based synthesis, GANalyzer can manipulate a randomly sampled latent vector such that it results in an image with specific facial attributes (\textit{e.g.} generating images of angry women).

Our proposed GANalyzer analyzes facial attributes of a wide range of synthesized images and their corresponding latent vector to recognize and interpret their relationship. Thus, we generate around $100K$ images using StyleGANs~\cite{karras2020training, karras2021alias} family as our training set. For any image in the training set, we utilize a set of different off-the-shelves classifiers to predict the corresponding facial attribute classes and label each image. More specifically, we use 4 different classifiers to predict image facial attributes including facial expression, gender, age, and race of each image in our training set. For each class (\textit{e.g} \textit{Happy} class from facial expression), we use the statistical variance of the covariance matrix of Eigenvectors of the latent vectors and the mean latent vector of that class to determine the relationship between the latent vectors and the specific facial attribute. Accordingly, we provide a transformation function for facial attribute editing and feature-based synthesis. We define our proposed transformation function $\Psi(z, S)$ where $z$ is the latent vector and $S$ is the target facial attribute we want to modify. As Fig.\ref{fig:transformation_function} shows, our proposed transformation method decomposes a latent vector $z$ into two vectors $z_{fb}$, and $z_{id}$. While $z_{fb}$ is designed to perform feature-based synthesis, $z_{id}$ is used for the facial attribute editing approach.

\begin{figure}[t]
  \centering
  \includegraphics[width=\columnwidth]{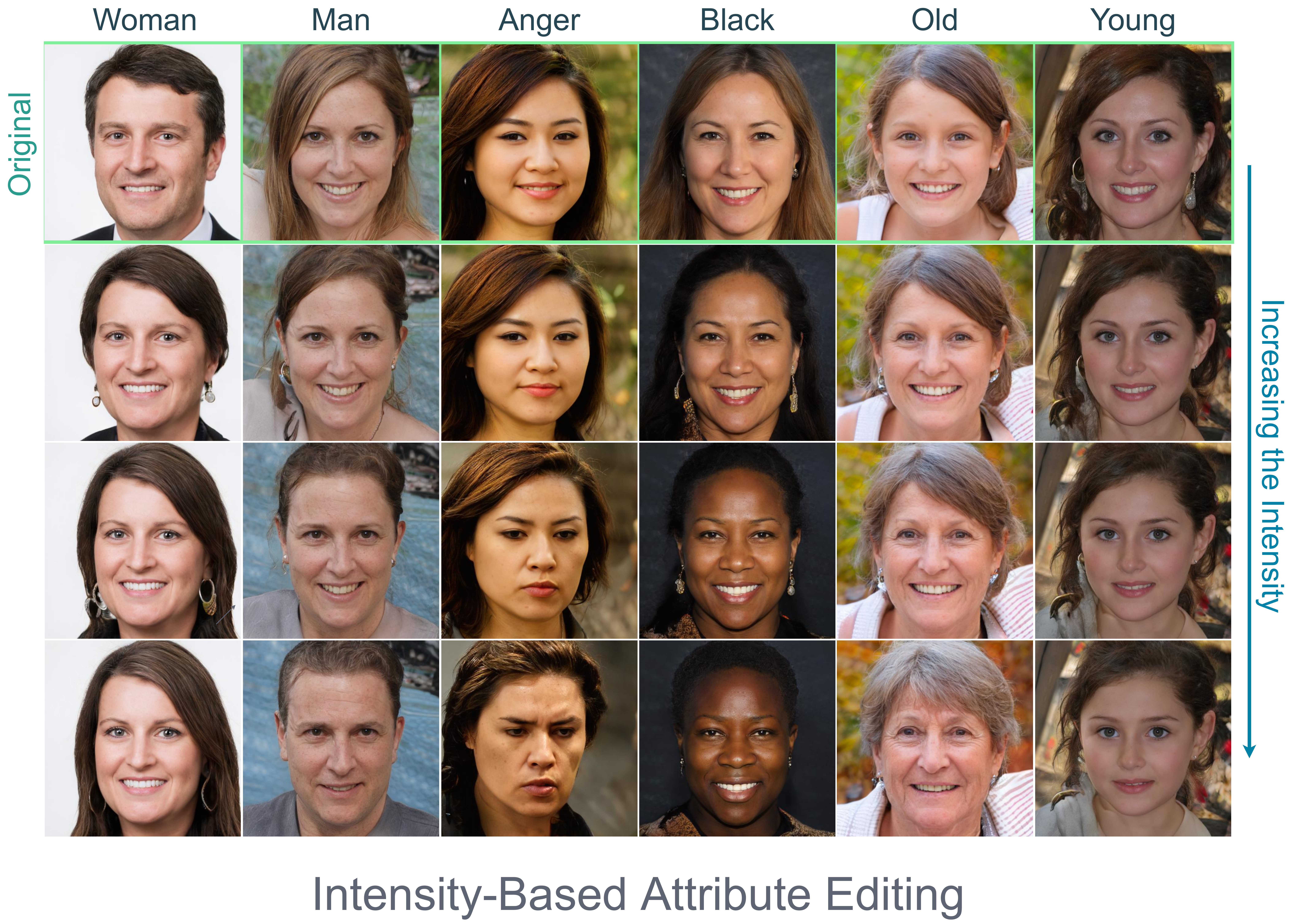}
  \caption{ The figure shows the intensity-based facial attribute editing using $\Psi$.}
  \label{fig:intensity_based_FAE}
\end{figure}

Moreover, for both facial attribute editing and feature-based image synthesis, our proposed GANalyzer has control over the intensity of the desired target facial attribute. In other words, not can only GANalyzer modify the facial attributes of a synthesized image, but it also can control how strong or weak we want such facial attributes to be presented in the synthesized face. To illustrate, say we want to generate a face that is prototypically Black. We can vary how Black or White that face appears (e.g. changes to skin tone, hairstyle, and face shape) by increasing or decreasing that dimension, respectively. Fig.~\ref{fig:intensity_based_FAE} shows a few examples of the intensity-based facial attribute editing provided by our proposed GANalyzer. 

In addition, our proposed GANalyzer can be extended from \textit{single} facial attribute manipulation to \textit{multiple} attribute manipulation, where we manipulate the latent vector to modify more than one facial attribute in the corresponding generated image. Fig.~\ref{fig:multiple_facial_attribute_editing} shows a few examples of multiple facial attribute editing. Likewise, GANalyzer can perform multiple feature-based syntheses too, where we can generate images with multiple desired facial attributes (\textit{e.g.} a Black woman). 

The contribution of the paper can be highlighted as the followings:
\begin{itemize}
    \item We propose a method for interpreting, and analyzing the latent space of GANs designed for human face synthesis. 
    \item We propose a transformation function called $\Psi$ for single and multiple feature-based syntheses as well as facial attribute editing. 
    \item Our proposed method has control over the intensity of the target facial attributes. 
    \item Using our feature-based image synthesis, we generate a facial expression database having \textit{Happy}, \textit{Neutral}, and \textit{Angry} emotions, with improved diversity with respect to perceived Age, Gender, and Racial Prototypicality. The dataset will be publicly available for research purposes.
\end{itemize}

The remainder of this paper is organized as follows. Sec.~\ref{sec_related_work} reviews the related work in GANs latent space analysis. Sec.~\ref{sec_methodology} describes our proposed methodology for the interpretation of GANs latent space and our transformation method for feature-based synthesis and facial attribute editing. Sec.~\ref{sec_evaluation} provides the experimental results, and finally, Sec.~\ref{sec_conclusion} concludes the paper with some discussions on the proposed method and future research directions.

\section{Related Work}\label{sec_related_work}
\textit{Generative Adversarial Networks:} GANs, first introduced by Goodfellow~\etal~\cite{goodfellow2020generative}, are among the most powerful methods for photo-realistic image synthesis. The input of GANs is randomly sampled latent vectors from a known distribution (most commonly a Gaussian distribution). During an adversarial-based training process, GANs learn how to convert the input noise vector (\textit{a.k.a.} latent vector) to the distribution of output data. Many variations have been proposed to improve the synthesis quality and make the training process stable~\cite{karras2017progressive, karras2019style, karras2020analyzing, arjovsky2017wasserstein, gulrajani2017improved, berthelot2017began, miyato2018spectral, zhang2019self, brock2018large, petzka2017regularization, yaz2018unusual}. Despite the variety of applications including image editing~\cite{lample2017fader, bau2020semantic, zhu2020domain, cheng2020sequential, cherepkov2021navigating}, image inpainting~\cite{suraj2021deep, yeh2017semantic, yu2019free, hedjazi2021efficient, yuan2019image}, super resolution~\cite{ledig2017photo, wang2018esrgan, zhang2019ranksrgan, rakotonirina2020esrgan+, mahapatra2019image}, video synthesis~\cite{wang2018video, wang2019few, mallya2020world}, \textit{etc.}, there is little work on the analysis and the interpretation of the latent space which can eventually highlight how the modification of a latent vector can affect the synthesized output.

\textit{Latent Space Interpretation and Analysis:} Latent space of GANs can be taken as a Riemannian manifold~\cite{chen2018metrics, arvanitidis2017latent}. Thus, interpolation in the latent space~\cite{laine2018feature, shao2018riemannian} has been studied to make the output image vary smoothly from a source image to a target image.  

Modifying the training process to learn interpretable factorized representation is among the methods for GANs latent space analysis~\cite{chen2016infogan, higgins2016beta}. Chen~\etal~\cite{chen2016infogan} proposed InfoGAN which can learn disentangled representations by maximizing the mutual information between a small subset of the latent variables and the observation. Li~\etal~\cite{li2021surrogate} used an auxiliary mapping network to model the relationship between latent vectors and the predicted semantic score of the corresponding generated images.

Vector arithmetic applied to the latent space can semantically manipulate the generated images~\cite{radford2015unsupervised, upchurch2017deep}. Vector arithmetic is model agnostic, and it can be categorized as supervised~\cite{plumerault2020controlling, shen2020interpreting, shen2020interfacegan}, and unsupervised~\cite{voynov2020unsupervised, harkonen2020ganspace, shen2021closed} methods. Supervised methods use a set of classifiers to label the properties of the generated images, and accordingly manipulate the latent vectors. Shen~\etal~\cite{shen2020interpreting} trained a linear Support Vector Machine (SVM) on latent vectors to find a decision hyperplane. More recently, InterfaceGAN~\cite{shen2020interfacegan} proposed how to learn a hyper-plane for binary classification in the latent space for each facial semantic, and ultimately use interpolation for modifying the attribute of the generated images. Plumerault~\etal~\cite{plumerault2020controlling} proposed a method to advance the interpretability of the latent space which controls specific properties of the generated image like the position or scale of the object in the image. Voynov~\etal~\cite{voynov2020unsupervised} proposed an unsupervised method to manipulate the latent vector by finding the directions corresponding to sensible semantics. Harkonen~\etal~\cite{harkonen2020ganspace} applied Principal Component Analysis on the latent space and proposed to control the semantics by layer-wise perturbation along the principal directions. Shen~\etal~\cite{shen2021closed} proposed a factorization algorithm for latent semantic discovery using pre-trained weights decomposition.

While unsupervised methods do not require different classifiers, supervised methods would provide more control over the manipulation of a specific facial attribute. From feature-based image synthesis, it is crucial to modify latent vectors such that generated images inherit the desired facial attribute. Thus, we proposed GANalyzer, following the supervised category. Moreover, GANalyzer is model agnostic and it can be applied over the latent space of a well-trained GAN.


%
\begin{figure}[t]
  \centering
  \includegraphics[width=\columnwidth]{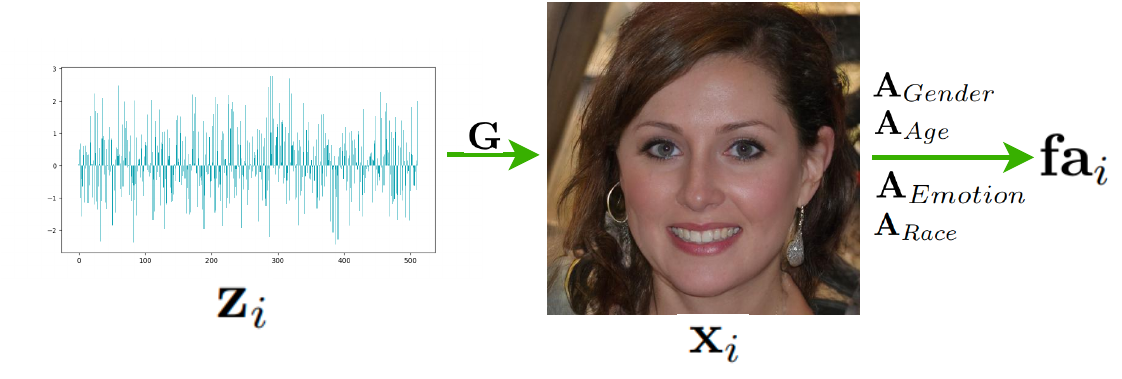}
  \caption{The process of generating an image $x_i$, from a random noise $z_i$, and its corresponding facial attribute set ${fa}_{i}$, created using pre-trained classifiers.}
  \label{fig:semantic_definition_process}
\end{figure}

\section{Methodology}\label{sec_methodology}
In this section, we first introduce facial attribute recognition and the labeling process of images in our training set. Then, we provide a deep analysis of the latent space of well-trained GANs (\textit{e.x.} StyleGANs~\cite{karras2020training, karras2021alias} family), and consequently, introduce our proposed transformation function. Afterward, we extend the transformation function for multiple facial attribute editing and feature-based synthesis. Finally, we analyze the entanglement between different facial attributes and features and provide a solution for disentangled facial attribute editing and feature-based synthesis.

\subsection{Latent Space and Attribute Recognition}
We can formulate a GAN as a function $\mathbf{G}:~\mathcal{Z}~\to~\mathcal{X}$, where $\mathcal{Z}$ is the latent space, and $\mathcal{X}$ represents the image space. Most of the previously proposed GANs~\cite{karras2019style} sample $\mathcal{Z}~\subseteq~\mathbb{R}^{d}$, from a Gaussian distribution $N(0,~\mathds{1}_d)$, where $d$ defines the dimensions of the latent space. Since $\mathbf{G}$ is a deterministic function, for any randomly sampled latent vector $\mathbf{z}~\in~\mathcal{Z}$, there exists a unique image $\mathbf{x}~\in~\mathcal{X}$. For any image $\mathbf{x}$, we can calculate a set of different facial attributes (\textit{e.g.} perceived facial expression, gender, age, race,\textit{ etc.}), using off-the-shelf classifiers. Hence, we can define an arbitrary number of facial attribute recognition functions $\mathbf{A}_c:~\mathbf{x}~\to~\mathbf{c}$, where $\mathbf{A}_c$ annotate a specific facial attribute of $\mathbf{x}$. Moreover, $\mathbf{c}~\subseteq~\mathbb{R}^{m}$ is a probability vector, and $m$ defines number of the classes in $\mathbf{A}_c$ (\textit{e.x.} \textit{women} or \textit{men} from gender class).  

In this paper, we choose to define four facial attribute recognition functions including \textit{facial expression}, \textit{gender}, \textit{age}, and \textit{race}, as follows in Eq.~\ref{eq:semantic_score_functions}:
\begin{align}\label{eq:semantic_score_functions}
    \begin{matrix*}[l]
    \mathbf{A}_{Gender}:\mathbf{x} \to \mathbf{gnd} & s.t & \mathbf{gnd} \subseteq \mathbb{R}^{2} \in \begin{adjustbox}{width=45pt}$\{\text{woman, man}\}$ \end{adjustbox} \\
    \mathbf{A}_{Age}: \mathbf{x} \to \mathbf{age} & s.t & \mathbf{age}  \subseteq  \mathbb{R}^{2} \in \begin{adjustbox}{width=40pt}$\{\text{Young, Old}\}$ \end{adjustbox} \\
    \mathbf{A}_{Emotion}: \mathbf{x} \to \mathbf{emo} & s.t & \mathbf{emo} \subseteq \mathbb{R}^{3} \in \begin{adjustbox}{width=55pt}$\{\text{Happy, Neutral, Angry}\}$ \end{adjustbox} \\
    \mathbf{A}_{Race}:~\mathbf{x} \to \mathbf{rce}  & s.t  & \mathbf{rce} \subseteq  \mathbb{R}^{2} \in \begin{adjustbox}{width=60pt}$\{\text{Black, White, Others}\}$ \end{adjustbox}
    \end{matrix*}
\end{align}

As mentioned above, we use pre-trained classifiers to label the synthesized images. For $\mathbf{A}_{Gender}$, $\mathbf{A}_{Age}$ and $\mathbf{A}_{Race}$, and $\mathbf{A}_{Emotion}$, we use the classifier proposed by Rothe~\etal~\cite{Rothe-IJCV-2018}, Serengil~\etal~\cite{serengil2021lightface}, and Fard~\etal~\cite{fard2022ad}, respectively. Needless to say, it is possible to extend the modification dimensionality by utilizing other pre-trained classifiers or recognition methods (\textit{e.x.} facial landmark and head pose estimators~\cite{fard2021asmnet, fard2022facial, fard2022acr, fard2022sagittal} if modification of facial pose of the synthesized images is needed.)
While the output of the original classifiers might be different than the output of our facial attribute recognition functions, we only use the probability scores assigned to the class of our interest, and ignore the rest. To illustrate, for facial expression recognition, the classifier proposed by Fard~\etal~\cite{fard2022ad} is designed to predict a 7-dimensional probability vector representing the probability of the following facial expressions: neutral, happy, sad, surprise, fear, disgust, and anger. However, to make the facial emotion space simpler, and more comparable to the other facial attributes, we define our $\mathbf{A}_{Emotion}$ to only consider the facial expressions we are interested in manipulating: happy, neutral, and angry. Likewise, we simplified race, too, by focusing on just three racial groups which are Black, White, and Others: \textit{Indian}, \textit{Middle-eastern}, and \textit{Latinx}.

Next, we synthesize $N$ number of images using the generative function, $\mathbf{G}$. As Fig.~\ref{fig:semantic_definition_process} shows, for any image $\mathbf{x}_i$ generated by the corresponding latent vector $\mathbf{z}_i$, we calculate its corresponding facial attributes set, $\mathbf{fa}_i$, as follows in Eq.~\ref{eq:sem_set_0}:
\begin{align}\label{eq:sem_set_0}
\mathbf{fa}_i = \begin{adjustbox}{width=180pt}$
                    \{ \mathbf{A}_{Gender}(\mathbf{x}_i),
                    \mathbf{A}_{Age}(\mathbf{x}_i),
                    \mathbf{A}_{Emotion}(\mathbf{x}_i),
                    \mathbf{A}_{Race}(\mathbf{x}_i)
                    \}$ \end{adjustbox}
\end{align}
Then, using Eq.\ref{eq:semantic_score_functions} we can write down $\mathbf{fa}_i $ as follows:
\begin{align}\label{eq:sem_set_1}
\mathbf{fa}_i = \{  \mathbf{gnd}_i,
                        \mathbf{age}_i,
                        \mathbf{emo}_i,
                        \mathbf{rce}_i
                    \} \subseteq  \mathbb{R}^{10}
\end{align}
After creating the corresponding facial attributes set for all the synthesized images, we use the similarity between the latent vectors that are categorized in the same class to propose our transformation function $\Psi$.  For instance, if we want to model the \textit{Anger} facial attribute from the facial expression class, we use the similarity between the corresponding latent vectors of the images which are labeled as \textit{Angry} to create our transformation function.

\subsection{Latent Space Analysis \& Transformation Functions}\label{sec_latent_space_manipulation}
We design our proposed GANalyzer framework to model any of the facial attributes in Eq.~\ref{eq:semantic_score_functions}, by introducing a unique transformation function $\Psi$ with respect to each facial attribute. For any arbitrary facial attribute object, $fa\_obj \in \mathbf{fa}$ (\textit{e.g.} anger from emotion), we define the transformation function in Eq.~\ref{eq:transformation_function} as follows:
\begin{align}\label{eq:transformation_function}
\Psi(\mathbf{z}_i~|~fa\_obj) ~\to ~ \mathbf{z}^{fb}_{i},~\mathbf{z}^{id}_i
\end{align}
Our proposed transformation function $\Psi$ manipulates an input latent vector $\mathbf{z}_i$, and creates 2 output latent vectors $\mathbf{z}^{fb}_{i}$, and $\mathbf{z}^{id}_i$. Then, using the generative function $\mathbf{G}$, we synthesize $\mathbf{x}_i$, $\mathbf{x}^{fb}_i$, and $\mathbf{x}^{id}_i$ corresponding to the latent vectors $\mathbf{z}_i$, $\mathbf{z}^{fb}_{i}$, and $\mathbf{z}^{id}_i$ respectively. 

We use $\mathbf{z}^{id}_i$ for facial attribute editing, as we designed the transformation function to manipulate the input latent vector $\mathbf{z}_i$ such that the corresponding generated image, $\mathbf{x}^{id}_i$, inherits from the $fa\_obj$ facial attributes, while its identity is preserved and almost similar to the identity of $\mathbf{x}_i$. Likewise, $\mathbf{z}^{fb}_i$ is designed for feature-based synthesis, which is agnostic to the identity. Thus, the corresponding generated image $\mathbf{x}^{fb}_i$ inherits from the $fa\_obj$ facial attributes, while the identities of $\mathbf{x}^{fb}_i$ and $\mathbf{x}_i$ would be different from each other. 

To create a transformation function corresponding to the facial attribute object $fa\_obj$, we first create a set of latent vectors from our training set and call it $\mathcal{Z}_{fa\_obj} = \{ \mathbf{z}_0, \mathbf{z}_1, ..., \mathbf{z}_k\}$. For each latent vector $\mathbf{z}_i~\in~\mathcal{Z}_{fa\_obj}$, 
we have $fa\_obj~\in~\mathbf{fa}_i$, which means the target facial attribute $fa\_obj$ must be in the corresponding facial attributes set $\mathbf{fa}_i$. To clarify, we can define $fa\_obj$ as \textit{Anger} from the emotion class, and the facial attribute set $\mathcal{Z}_{anger}$, that includes the latent vectors corresponding to all the synthesized images labeled as \textit{Angry}. 

\begin{figure}[t]
  \centering
  \includegraphics[width=\columnwidth]{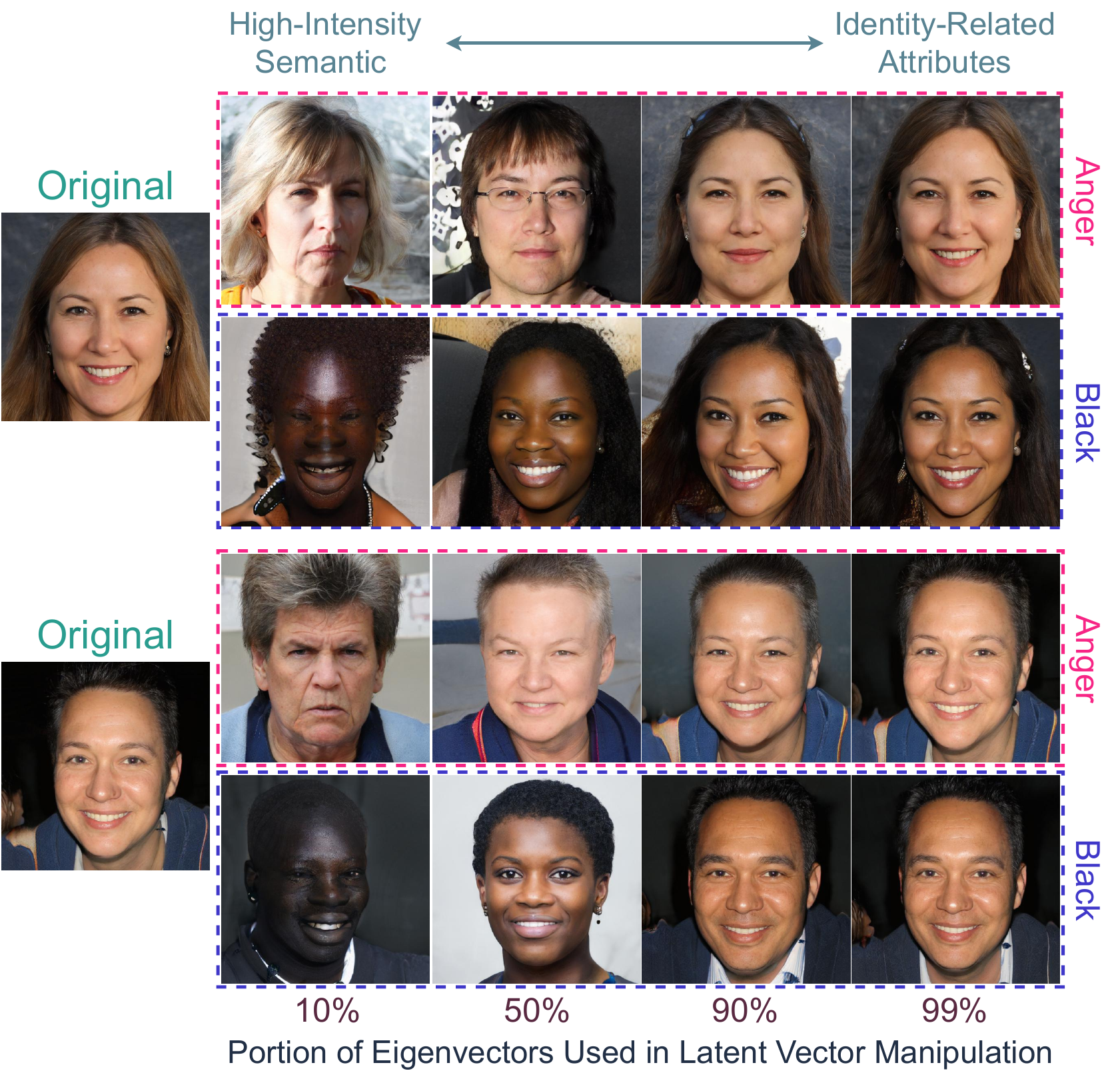}
  \caption{For feature-based synthesis, the portion of the Eigenvectors of the Covariance matrix defines the intensity of the target facial attribute in the generated image. As the figure shows, using $10\%$ of the Eigenvalues will result in an image with high-intensity facial attributes (Anger, and Black), while using $99\%$ of the Eigenvectors results in an image with a very low-intensity facial attribute, with the same identity compared to the original image.}
  \label{fig:eigenvectors_portion}
\end{figure}

Inspired by Cootes~\etal~\cite{cootes2000introduction, cootes2004statistical}, we use the Eigenvectors of the Covariance matrix corresponding to $\mathcal{Z}_{fa\_obj}$ to propose our transformation functions. Firstly, we define $\mathbf{V}_{fa\_obj}~=~\{v_1, v_2, ..., v_t\}$ as the set of all the Eigenvectors of the Covariance matrix of $\mathcal{Z}_{fa\_obj}$. Then, we define $\mathbf{m}_{fa\_obj}$ as the element-wise statistical mean vector of $\mathcal{Z}_{fa\_obj}$. Finally, we define $\mathbf{b}_{fa\_obj}$ vector as the following in Eq.~\ref{eq:b_vector}:
\begin{align}\label{eq:b_vector}
\mathbf{b}_{fa\_obj_{(t \times 1})} = \mathbf{V}_{fa\_obj_{(t \times d})}^\intercal (\mathbf{z}_{i_{(d \times 1})} - \mathbf{m}_{fa\_obj_{(d \times 1})})
\end{align}
Considering that the $i^{th}$ element of the statistical variance (\textit{a.k.a.} Eigenvalues) corresponding to $\mathbf{V}_{fa\_obj}$ is $\lambda_i$. Following Cootes~\etal~\cite{cootes2000introduction, cootes2004statistical}, we limit $b_i$, the $i^{th}$ element of the vector $\mathbf{b}_{fa\_obj}$, such that $b_i\in[-3\sqrt{\lambda_i}, +3\sqrt{\lambda_i}]$, and create a new vector called $\mathbf{\Tilde{b}}$. We can estimate a latent vector $\mathbf{z}_i~\in~\mathcal{Z}_{fa\_obj}$ using Eq.~\ref{eq:latent_vector_estimation} as follows:
\begin{align}\label{eq:latent_vector_estimation}
 \mathbf{\Tilde{z}}_i \approx \mathbf{m}_{fa\_obj} + \mathbf{V}_{fa\_obj}  \mathbf{\Tilde{b}}
\end{align}
\begin{figure*}[t!]
  \centering
  \includegraphics[width=18cm]{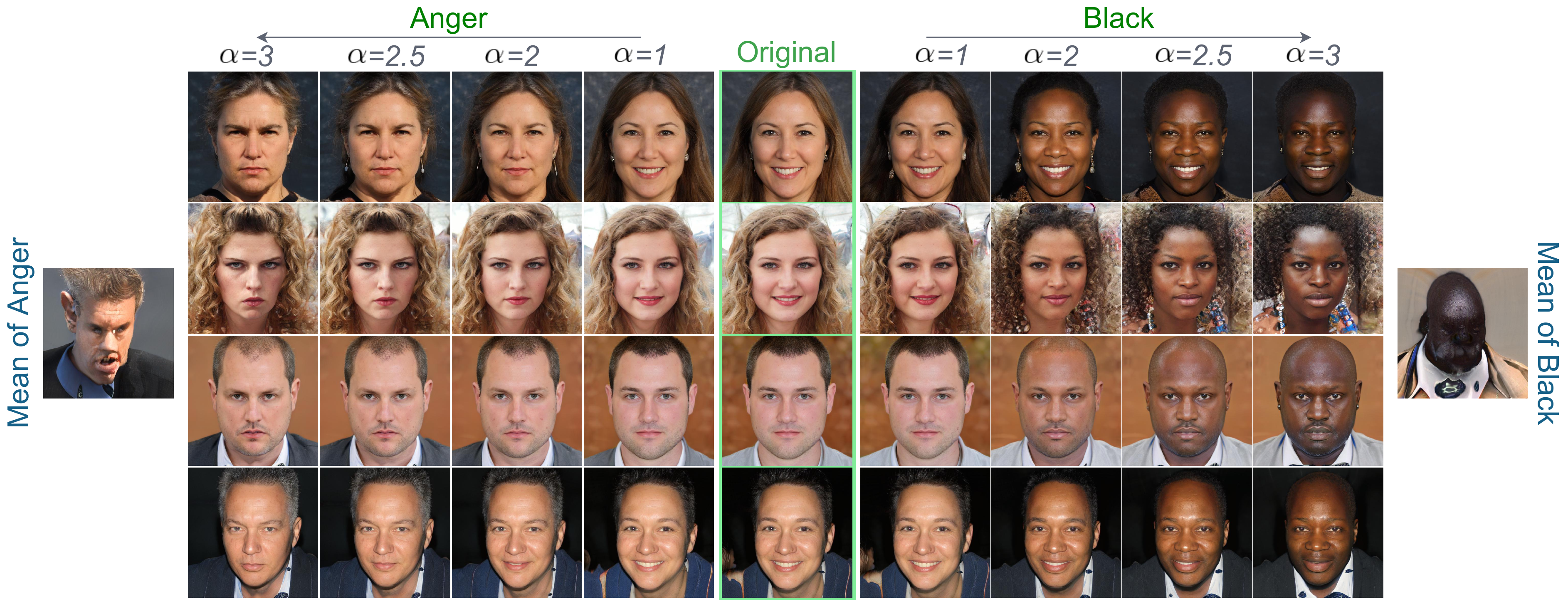}
  \caption{Facial attribute editing with respect to different values of $\alpha$. The greater the value of  $\alpha$, the higher would be the intensity of the target facial attribute. In fact, $\alpha$ highlights the effect of the corresponding mean vector in the manipulated latent vector.}
  \label{fig:identity_mean}
\end{figure*}

We use Eq.~\ref{eq:latent_vector_estimation} as the building block for defining our transformation function $\Psi$. We defined $\mathbf{V}_{fa\_obj}$ as the set of Eigenvectors of the Covariance matrix of the latent vectors having a \textit{common} facial attribute ${fa\_obj}$. Thus, the first Eigenvectors represent the features and attributes which model ${fa\_obj}$ facial attribute. To clarify, if we want to model \textit{Anger} (and thus we define our ${fa\_obj}$ as Anger) using Eq.~\ref{eq:latent_vector_estimation}, the first (\textit{a.k.a.} the most important) Eigenvectors 
in $\mathbf{V}_{Anger}$ models the \textit{Anger}-related features. In other words, to model a low-frequency facial attribute ${fa\_obj}$ (such as perceived facial emotion, race, \textit{etc.}) in a dataset, we can use the most important Eigenvectors of the corresponding Covariance matrix of a subset where all its samples have ${fa\_obj}$ (\textit{e.g. a subset where all samples are classified as happy}). Likewise, the high-frequency features (sample-specific features, which are mostly identity-related features) can be modeled by the least important Eigenvectors. Later, in Sec.~\ref{sec_tf_SB}, we define a hyper-parameter $\beta \in [0, 100]$, indicating the portion of the first Eigenvectors that are used to create the transformation function.

In Fig.~\ref{fig:eigenvectors_portion}, we demonstrate the effect of the Eigenvectors in image synthesis. Accordingly, using $10\%$ of the most important Eigenvectors for latent vector manipulation, using Eq.~\ref{eq:latent_vector_estimation}, results in an image that intensively inherits from the corresponding facial attribute. However, regarding the identity comparison, the original image and the modified target image hardly have any identity-related similarities. As we gradually choose a larger portion of the Eigenvectors, we observe a reduction in the intensity of the corresponding features (\textit{e.x.} \textit{Anger} in Fig.~\ref{fig:eigenvectors_portion}), while the identity-related feature increases.


\subsubsection{Facial Attribute Editing}\label{sec_tf_FAE}
For facial attribute editing, the identity of the generated image after manipulation must be relatively similar to the source image. As discussed in Sec.~\ref{sec_latent_space_manipulation}, using Eq.~\ref{eq:latent_vector_estimation}, we need to use almost all the Eigenvectors of the Covariance matrix such that both low-frequency and high-frequency features of the source and the target images become relatively similar. However, using all Eigenvectors, makes the generated latent vector $\mathbf{\Tilde{z}}_i$ to be relatively similar to the source vector $\mathbf{z}_i$. As Fig.\ref{fig:eigenvectors_portion} shows, using either $90\%$ or $99\%$ of the Eigenvectors, the target synthesized image could be relatively similar to the original image, while there is no guarantee that the target image inherits from $fa\_obj$. 

\begin{figure}[t]
  \centering
  \includegraphics[width=\columnwidth]{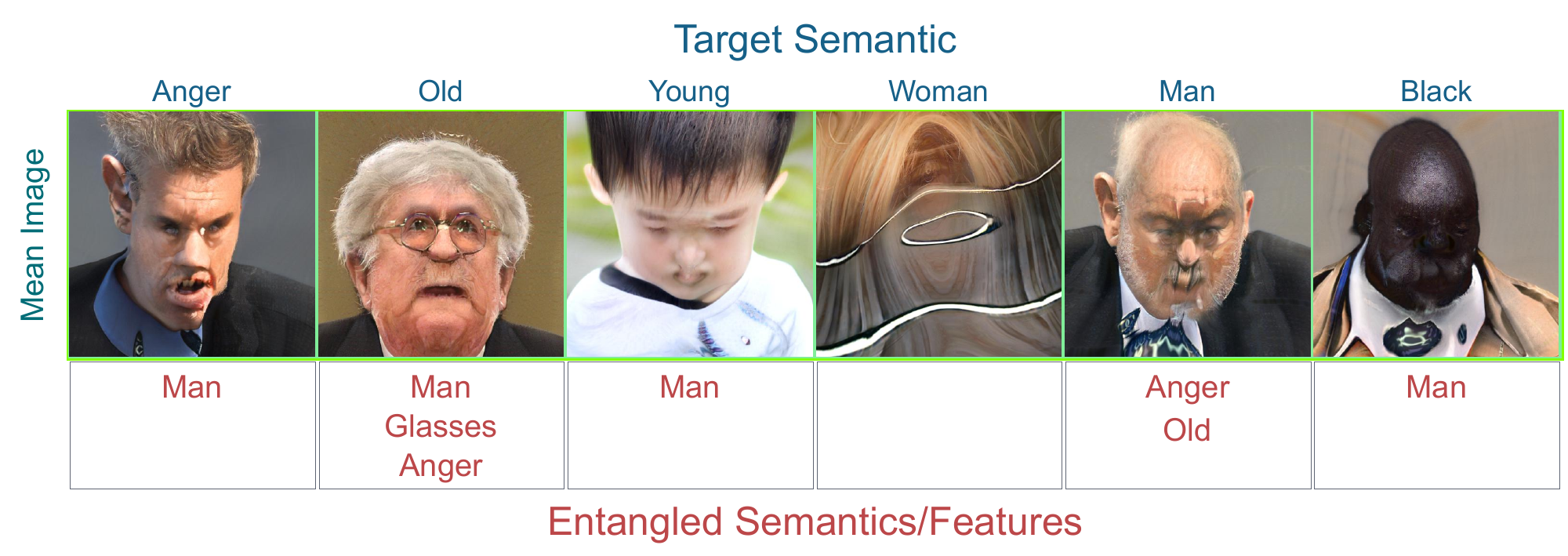}
  \caption{The generated image from the mean latent vector of different facial attributes shows the entanglement between the facial attributes. As an example, \textit{Anger} is entangled with \textit{Man}. }
  \label{fig:mean_entanglement}
\end{figure}

To overcome this issue, we propose to increase the impact of the mean latent vector in Eq.~\ref{eq:transformation_function}. The mean latent vector $\mathbf{m}_{fa\_obj}$, created from the element-wise average of all the latent vectors having $fa\_obj$. Hence, as Fig.~\ref{fig:mean_entanglement} shows, the images generated from the mean latent vectors highly represent the corresponding facial attributes. Consequently, increasing the weight of the mean latent vector, $\mathbf{m}_{fa\_obj}$, while using all of the Eigenvectors of the Covariance Matrix, in the estimation of the source latent vector $\mathbf{z}_i$ in Eq.~\ref{eq:transformation_function} results in an image which is relatively similar to the original image in terms of the identity, while it inherits from the desired $fa\_obj$ facial attribute. We propose Eq.~\ref{eq:latent_vector_estimation_identity} for facial attribute editing as follows:
\begin{align}\label{eq:latent_vector_estimation_identity}
 \mathbf{z}^{id}_{i} = \alpha ~ \mathbf{m}_{fa\_obj} + \mathbf{V}_{fa\_obj}  \mathbf{\Tilde{b}}
\end{align}
where $\alpha \geq 1 $ is a hyper-parameter added to Eq.~\ref{eq:latent_vector_estimation_identity} to intensify the effect of the mean vector $\mathbf{m}_{fa\_obj}$ in the creation of $\mathbf{z}^{id}_{i}$. As Fig.~\ref{fig:identity_mean} shows, the mean vector $\mathbf{m}_{fa\_obj}$ mostly preserves the most common facial attributes and features within the corresponding set of latent vectors having $fa\_obj$ facial attribute. Thus, considering $fa\_obj$ set, $\mathbf{G}(\mathbf{m}_{fa\_obj})$ is an image $\mathbf{x}_{m}$, which intensively inherits from the $fa\_obj$. Thus, adding weight to the mean vector $\mathbf{m}_{fa\_obj}$ in Eq.~\ref{eq:latent_vector_estimation_identity}, would modify the input latent vector $\mathbf{z}_{i}$ such that the generated latent vector $\mathbf{z}^{id}_{i}$ results in an image that inherits facial attributes from ${fa\_obj}$, while its identity-related features are similar to the image generated from $\mathbf{z}_{i}$.

\subsubsection{Feature-Base Synthesis}\label{sec_tf_SB}
For feature-based synthesis, instead of using the complete set of Eigenvectors of the Covariance matrix $\mathbf{V}_{fa\_obj}$, we introduce a hyper-parameter $\beta \in [0,100]$ indicating the portion of the first Eigenvectors. Based on the value of $\beta$, we define $\mathbf{\Tilde{V}}_{fa\_obj}$, which is a subset of $\mathbf{V}_{fa\_obj}$, having only the first $\beta$ portion of the Eigenvectors. Then, we define Eq.~\ref{eq:latent_vector_estimation_semantic} for feature-based manipulation:
\begin{align}\label{eq:latent_vector_estimation_semantic}
 \mathbf{z}^{fb}_{i} = \mathbf{m}_{fa\_obj} + \mathbf{\Tilde{V}}_{fa\_obj} \mathbf{\Tilde{b}}
\end{align}

Using Equations \ref{eq:transformation_function}, \ref{eq:latent_vector_estimation_identity} and \ref{eq:latent_vector_estimation_semantic}, we propose our transformation function $\Psi$ in Eq.~\ref{eq:transformation_function_d} as follows:
\begin{align}\label{eq:transformation_function_d}
    \begin{matrix*}[l]
        \Psi(\mathbf{z}_i~|~fa\_obj) ~\to ~ \mathbf{z}^{fb}_{i},~\mathbf{z}^{id}_i \\
        \mathbf{z}^{id}_{i} = \alpha ~ \mathbf{m}_{fa\_obj} + \mathbf{V}_{fa\_obj}  \mathbf{\Tilde{b}} \\
        \mathbf{z}^{fb}_{i} = \mathbf{m}_{fa\_obj} + \mathbf{\Tilde{V}}_{fa\_obj} \mathbf{\Tilde{b}}
    \end{matrix*}
\end{align}
The value of $\alpha$ needs to be selected based on how intensely we want to add $fa\_obj$ to the input image (see Fig.~\ref{fig:identity_mean}). Likewise, the parameter $\beta$ defines how intense we need the facial attribute $fa\_obj$ in the synthesized image. 
\begin{figure}[t]
  \centering
  \includegraphics[width=\columnwidth]{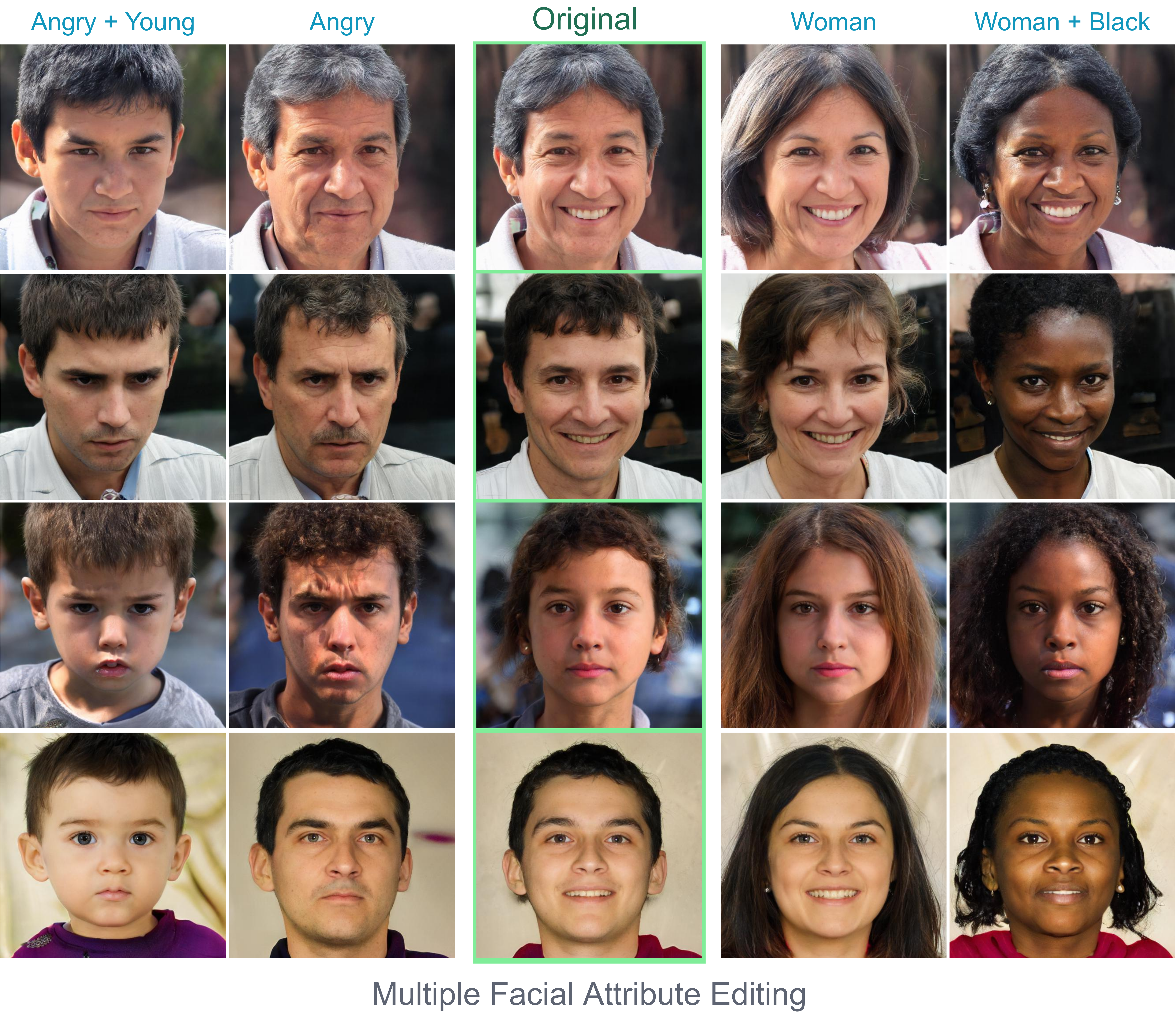}
  \caption{ Our proposed transformation function $\Psi$ can perform multiple facial attribute editing. As the figure shows, in the first column, we modified the original image to be \textit{Angry} and \textit{Young}, while in the last column, we modified the original image to be \textit{Woman} and \textit{Black}.}
  \label{fig:multiple_facial_attribute_editing}
\end{figure}

\subsection{Multiple Facial Attribute Manipulation}\label{sec_multiple_Semantic_Manipulation}
Multiple facial attribute manipulation is a useful tool for both facial attribute editing and feature-based synthesis. We propose a linear combination of a set of desired target facial attributes to perform these tasks. 

 Let's assume we want to manipulate a randomly sampled latent vector $\mathbf{z}_s$ such that the corresponding image generated from the manipulated latent vector $\mathbf{z}_d$ inherits from a set of facial attributes $fa\_set:~\{ fa\_obj_1, fa\_obj_2, ..., fa\_obj_n\}$. We select one of the elements of $fa\_set$ arbitrarily as the \textit{base} facial attribute and call it $fa\_objBase$. 

\textbf{Multiple Facial Attribute Editing}: For facial attribute editing, it is crucial to keep the identity of the original image. Hence, we use all of the Eigenvectors of the Covariance Matrix $\mathbf{V}_{fa\_objBase}$, corresponding to $fa\_objBase$ to keep both high- and low-frequency features and attributes of the original image after manipulation (see Sec.~\ref{sec_tf_FAE}). Then, following the method proposed for facial attribute editing, we use a linear combination of the mean latent vectors corresponding to each element of $fa\_set$. This ensures the synthesized image corresponding to the generated latent vector $\mathbf{z}_{mfa}$ inherits from all of the desired facial attributes. Finally, we propose Eq.~\ref{eq:mul_fa_based} as follows for multiple facial attribute editing:
\begin{align}\label{eq:mul_fa_based}
     \mathbf{z}_{mfa} =\sum_{1}^{n} \gamma_i \mathbf{m}_{fa\_obj_i} + 
     \mathbf{V}_{fa\_objBase}\mathbf{\Tilde{b}} 
\end{align}
where $n$ is the length of the set of facial attributes $fa\_set$, and $\gamma_i$ is a hyper-parameter that defines the intensity of the corresponding facial attribute in the image synthesized from $ \mathbf{z}_{mfa}$. As Fig~\ref{fig:multiple_facial_attribute_editing} shows, our proposed method can manipulate multiple facial attributes, while we still have control over the intensity level of the desired facial attributes as mentioned in Sec.~\ref{sec_latent_space_manipulation}.

\begin{figure}[t]
  \centering
  \includegraphics[width=\columnwidth]{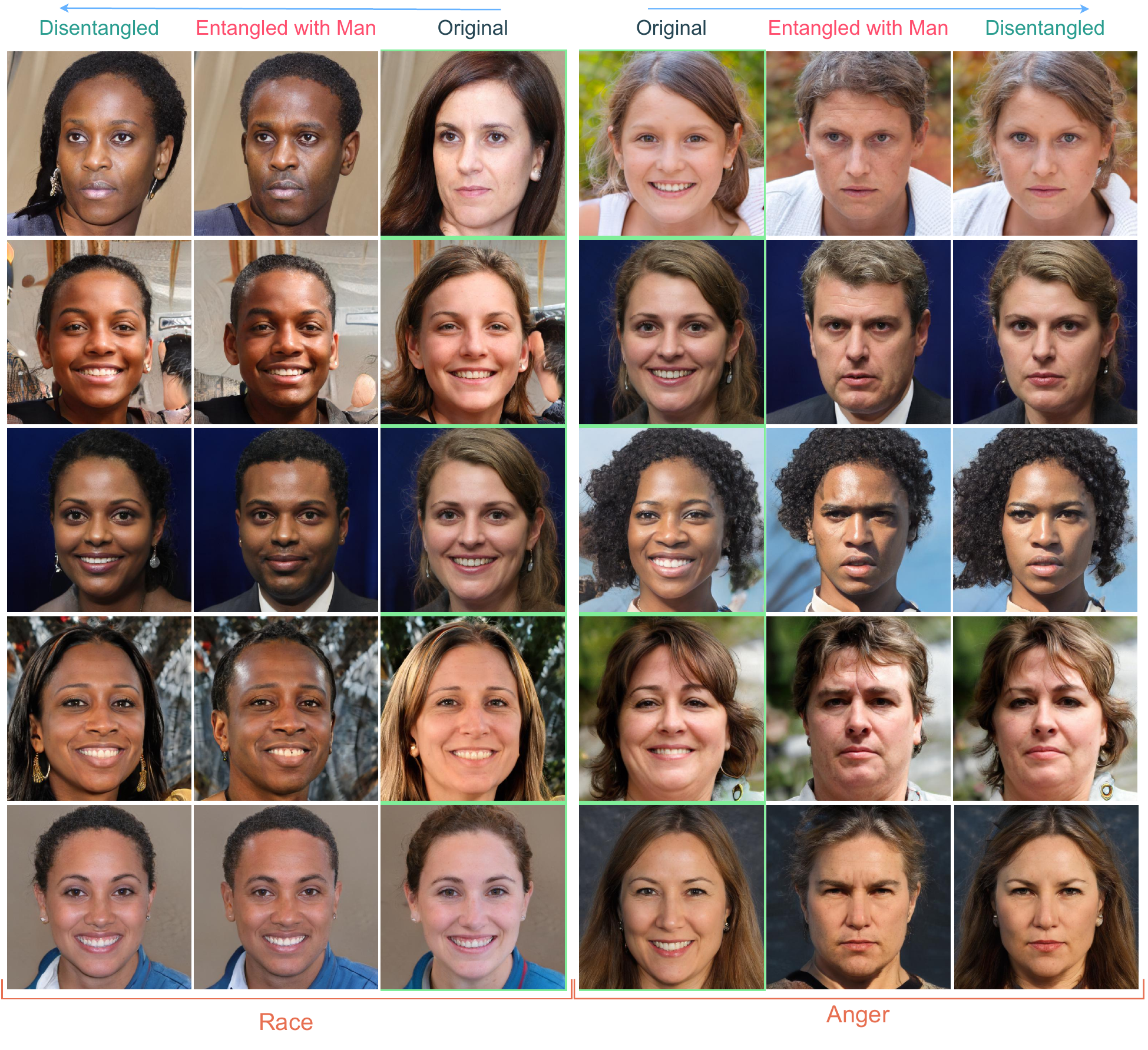}
  \caption{The figure shows the entanglement between some specific facial attributes, and how our proposed transformation function $\Psi$ can be modified to make these facial attributes disentangle from each other. As shown, \textit{Man} facial attribute is entangled with both \textit{Black}, and \textit{Anger}, and with a small modification to $\Psi$, all these facial attributes became disentangled.}
  \label{fig:disentanglement_samples}
\end{figure}

\textbf{Multiple Feature-Based Synthesis}: We follow the method proposed in Sec.~\ref{sec_tf_SB} for feature-based synthesis. Using its corresponding $\beta$, called $\beta_{Base}$, mean latent vector $\mathbf{m}_{fa\_objBase}$, and the subset of the Eigenvectors of the Covariance matrix $\mathbf{\Tilde{V}}_{fa\_objBase}$ (see Sec.~\ref{sec_tf_SB}, and Eq.~\ref{eq:latent_vector_estimation_semantic}), we introduce a transformation function for multiple feature-based syntheses as follows in Eq.~\ref{eq:mul_sem_based}:
\begin{align}\label{eq:mul_sem_based}
     \mathbf{z}_{mfe} =\sum_{1}^{n} \lambda_i \mathbf{m}_{fa\_obj_i} + 
     \mathbf{\Tilde{V}}_{fa\_objBase} \mathbf{\Tilde{b}} 
\end{align}
where $n$ is the length of the set of facial attributes $fa\_set$, and $\lambda_i$ is a hyper-parameter that defines the intensity of the corresponding facial attribute in the image synthesized from $\mathbf{z}_{mfe}$.

\begin{figure*}[t!]
  \centering
  \includegraphics[width=18cm]{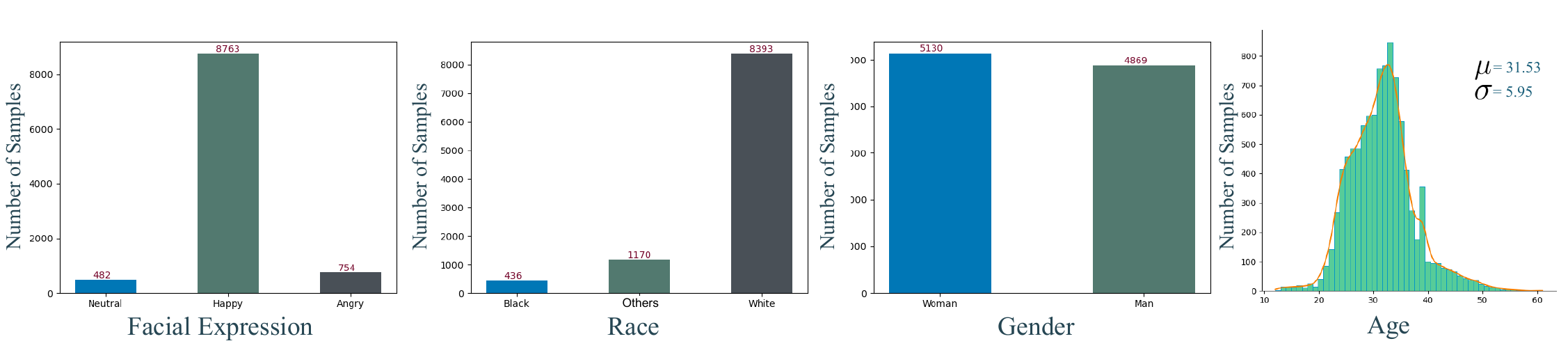}
  \caption{The distribution of the facial expression, race, gender, and age of the validation set.}
  \label{fig:evalset_histogram}
\end{figure*}

Our proposed method for multiple feature-based syntheses manipulates a randomly sampled latent vector, using the subset of the Eigenvectors of the Covariance matrix of one arbitrary facial attribute and a linear combination of the mean vectors of all of the desired facial attributes, to make sure that the generated latent vectors inherit from all of the desired facial attributes. We use this method to create a diverse dataset in Sec.~\ref{sec_proposed_dataset}.

\subsection{Attribute Entanglement Analysis}\label{sec_Semantic_Entanglement_and_Solution}
The correlation between different facial attributes of samples on which GANs are trained heavily relies on entanglement and disentanglement of the latent space~\cite{shen2020interfacegan, shen2020interpreting}. As an example, GANs usually entangle \textit{age} with \textit{glasses}~\cite{karras2019style, karras2020analyzing, karras2017progressive}. Entanglement in the latent space negatively affects both facial attribute editing and feature-based synthesis manipulation. For facial attribute editing, it is crucial to modify one desired facial attribute while keeping the others as well as the identity with no change. 

We propose an effective method for the investigation of the entanglement between different facial attributes and then propose a method to adjust the transformation function $\Psi$, to deal with the entanglement issue. To study the entanglement between different facial attributes, we create the mean latent vector $\mathbf{m}_{fa\_obj}$ corresponding to a \textit{desired} facial attribute $fa\_obj$. Then, using the GAN generative function $\mathbf{G}$, we synthesize the mean image $\mathbf{x}_{m}$ from the mean latent vector $\mathbf{m}_{fa\_obj}$. Afterward, we can visually observe the \textit{undesired} changes in facial attributes and facial attributes of the mean image, which can be taken as the entangled facial attributes with our desired facial attribute $fa\_obj$.

In Fig.~\ref{fig:mean_entanglement}, we calculate the mean latent vector and depict the corresponding mean image for \textit{Anger}, \textit{Old}, \textit{Young}, \textit{Woman}, and \textit{Black} facial attributes. As Fig.~\ref{fig:mean_entanglement} shows, visual analysis of the synthesized mean images can easily disclose the high degree of entanglement between specific facial attributes. As an example, it is obvious that \textit{Anger} is highly entangled with \textit{Man}, \textit{Old} with \textit{Man}, \textit{Glasses} and \textit{Anger}, and \textit{Black} with \textit{Man}. Also, in the mean image regarding the \textit{Woman} facial attribute, we do not observe any specific pattern, indicating that \textit{Woman} does not have a high entanglement to other facial attributes. While our proposed method can easily disclose feature/facial attribute entanglement, in Sec.~\ref{sec_eval_DisentanglementAnalysis}, we statistically measure the entanglement degree between the facial attributes to support our proposed technique for visual analysis of entanglement.

We further can modify our proposed transformation function to deal with the entanglement between the facial attributes. Assume we want to modify a randomly sampled latent vector $\mathbf{z}_s$ and create $\mathbf{z}_d$ such that the generated image $\mathbf{x} = \mathbf{G}(\mathbf{z}_d)$ inherits from a \textit{desired} facial attribute $fa\_obj_D$, while there is an \textit{undesired} facial attribute $fa\_obj_U$, which is entangled with $fa\_obj_D$. Using Eq.~\ref{eq:transformation_function_disentangled}, we can modify $\mathbf{z}_s$, and generate $\mathbf{z'}_d$ such that the synthesized image $\mathbf{x'} = \mathbf{G}(\mathbf{z'}_d)$ \textit{only} inherits from $fa\_obj_D$. 
\begin{align}\label{eq:transformation_function_disentangled}
    \begin{matrix*}[l]
        \mathbf{z'}_d = (\alpha ~ \mathbf{m}_{fa\_obj_D} - \delta ~ \mathbf{m}_{fa\_obj_U})
        + \mathbf{V}_{fa\_obj_D}  \mathbf{\Tilde{b}} 
    \end{matrix*}
\end{align}

$\mathbf{m}_{fa\_obj_U}$ is the mean of the \textit{undesired} facial attribute $fa\_obj_U$, and $\delta > 0$ is a hyper-parameter that sets the intensity of the \textit{undesired} facial attribute. More clearly, using $\mathbf{m}_{fa\_obj_U}$ with a negative weight $- \delta$ reduces the impact of the undesired facial attribute in the generation of the latent vector $\mathbf{z'}_d$. As an example, as Fig.~\ref{fig:mean_entanglement} suggests, the \textit{Anger} facial attribute is entangled with the \textit{Man}. Thus, can generate latent vectors which inherits from \textit{Anger}, while it is disentangled from \textit{Man} as follows:
\begin{align}\label{eq:anger_disentangled}
    \begin{matrix*}[l]
        \mathbf{z'}_d = (\alpha ~ \mathbf{m}_{fa\_Anger} - \delta ~ \mathbf{m}_{fa\_man})
        + \mathbf{V}_{fa\_Anger}  \mathbf{\Tilde{b}} 
    \end{matrix*}
\end{align}
As we discussed in Sec.~\ref{sec_latent_space_manipulation}, the term $\mathbf{V}_{fa\_Anger}  \mathbf{\Tilde{b}}$ preserves the identity of the image, while $\alpha ~ \mathbf{m}_{fa\_Anger}$ intensifies the \textit{Anger} of the corresponding synthesized image. Using $ - \delta ~ \mathbf{m}_{fa\_man}$, will reduce the intensity of the \textit{undesired} facial attribute (\textit{Man}) in the resulting image. Fig.~\ref{fig:disentanglement_samples} shows the performance of our method for disentangled facial attribute editing. Moreover, in Sec.~\ref{sec_eval_DisentanglementAnalysis}, we show how our proposed technique results in disentangled facial attribute editing.

\begin{figure}[t]
  \centering
  \includegraphics[width=\columnwidth]{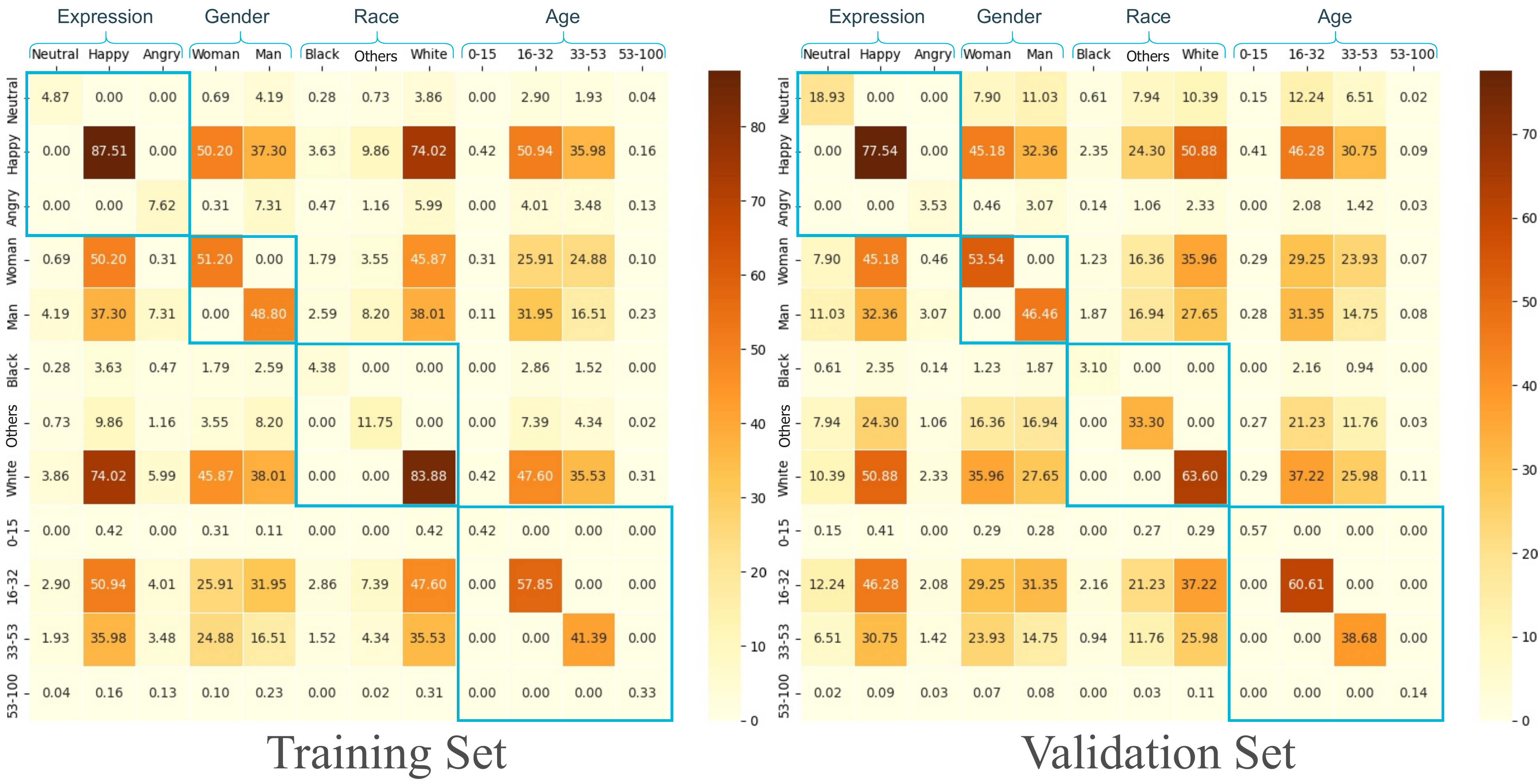}
  \caption{Covariance matrix of the training set and validation set. The figures show the portion of each facial attribute in the training and the validation set, as well as the correlation between different facial attributes. As an example, $77.54\% $of the samples in the training set, and $87.51\% $of the samples in the validation sets are classified as \textit{Happy}. For the training set, $45.18\%$ of this \textit{Happy} samples \textit{Woman}, and  $32.36\%$ are  \textit{Man}. Likewise, for the validation set, out of $87.51\%$ of this \textit{Happy} samples, regarding race, $3.63\%$ are labeled as \textit{Black}, $9.86\% as$ \textit{Others}, and$ 74.02\%$ as \textit{White}.}
  \label{fig:heatmap_histogram}
\end{figure}

\section{Evaluation}\label{sec_evaluation}
In this section, we first propose a method to evaluate our transformation function for facial attribute editing. Then, we evaluate the performance of feature-based synthesis. For the purpose of evaluation, we randomly generate 10K images using StyleGANs~\cite{karras2020training, karras2021alias} family and call it the \textit{validation set}. As mentioned in Sec.~\ref{sec_latent_space_manipulation}, for each image in our validation set, we use 4 pre-trained classifiers to measure the following facial attributes: Gender, Age, Facial Expression, and Race. For race, we use the pre-trained classifier provided by \cite{serengil2021lightface} for ethnicity prediction including \textit{Asian}, \textit{Indian}, \textit{Black}, \textit{White}, \textit{Middle-Eastern}, and \textit{Latinx}. As we are mostly interested in Black, and White in the context of this research for simplicity, and also to be able to interpret the relationship and the entanglement between different facial attributes more accurately. Hence, while we explicitly have \textit{White}, and \textit{Black}, we consider the \textit{Indian}, \textit{Middle-eastern}, and \textit{Latinx} as \textit{Others} race. 

Fig.~\ref{fig:evalset_histogram}, shows the facial attributes histogram of the validation set. Moreover, in Fig~\ref{fig:heatmap_histogram}, we show the Covariance matrix of both the training set and the evaluation set, indicating the portion of each facial attribute class, as well as the correlation between each facial attribute. We also use the Covariance matrix of the validation set in Sec.~\ref{sec_eval_DisentanglementAnalysis} for facial attribute entanglement analysis and to evaluate our proposed method for disentangled facial attribute editing.

\begin{figure}[t]
  \centering
  \includegraphics[width=\columnwidth]{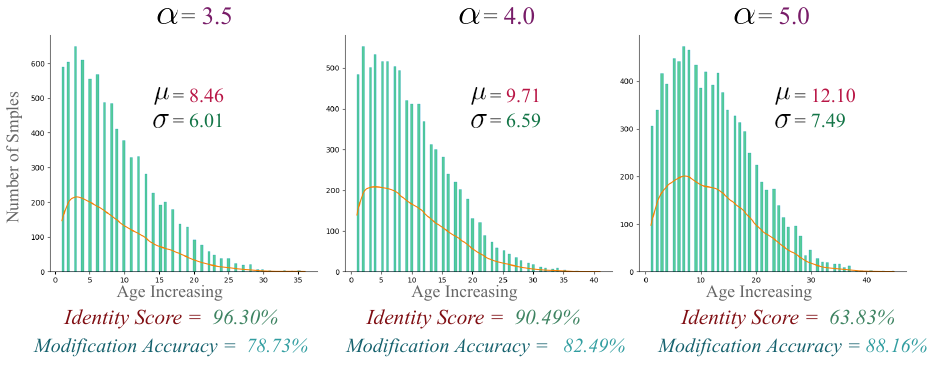}
  \caption{The evaluation of the transformation function for facial attribute editing with respect to \textit{Old} age. The figure shows the histogram of age differences, the modification accuracy, and the identity score for different values of $\alpha$. The orange chart shows the Kernel Density Estimation of the histogram.}
  \label{fig:to_old_chart_identity}
\end{figure}

\subsection{Facial Attribute Editing} \label{eval_Facial_Attribute_Editing}
To evaluate the performance of the transformation function in facial attribute editing, we conducted 5 different experiments. In each experiment, we define $fa\_obj$ as the target facial attribute that we want to modify, while we keep the identity of the synthesized image. Thus, for each image in the validation set, we modify the corresponding latent vector with respect to the target facial attribute $fa\_obj$ and then calculate the identity similarity score and the target facial attribute class. 

In each facial attribute modification experiment, we first modify the latent vector with respect to the target facial attribute and then, use the GAN generative function to generate the corresponding image. Next, we use RetinaFace~\cite{serengil2021lightface} to compare the identity of the original image with the modified synthesized image. We also use pre-trained classifiers to measure the target facial attributes of the synthesized images.

\begin{table}[b]
\caption{The evaluation of the transformation function for facial attribute editing with respect to \textit{Man} as well as \textit{Woman} gender.}
\label{tbl:identity_based_evaluation_fm}
\centering
\small
\resizebox{8cm}{!}{

\begin{tabular}{lccc}
\hline
                    & $\alpha= 2.5$     & $\alpha=3.0$  & $\alpha=4.0$     \\ \hline
Identity Score (to Man) (\%) & $79.71$             & $53.06$         & $41.89$         \\
Woman to Man (\%) & $94.39$             & $99.36$         & $99.60$        
\\ \hline
Identity Score (to Woman) (\%) &    $99.66$       &     $96.68$     &     $92.83$       \\
Man to Woman (\%) &    $66.48$       &     $92.39$     &     $97.09$    \\ \hline

\end{tabular}
}
\end{table}

In the first experiment, we evaluated our transformation function for editing the \textit{Woman} attribute. As Table~\ref{tbl:identity_based_evaluation_fm} shows, the identity and facial attribute scores heavily rely on the value of hyper-parameter $\alpha$ (see Eq.~\ref{eq:latent_vector_estimation_identity}). Setting $\alpha=2.5$, the identity score is $99.66 \%$, indicating that almost all of the modified images are identical to the original images, while the facial attribute score is $66.48\%$. As mentioned in Sec.~\ref{sec_latent_space_manipulation}, increasing the value of $\alpha$ to $3.5$ reduces the identity score to $96.68\%$, while the facial attribute score increases to $96.27\%$. Finally, we increased the value of $\alpha$ to $4.0$, and around $97.09\%$ of the images in the validation set labeled as \textit{Woman}, while the identity scores reduces to $92.83\%$.

\begin{figure}[t]
  \centering
  \includegraphics[width=\columnwidth]{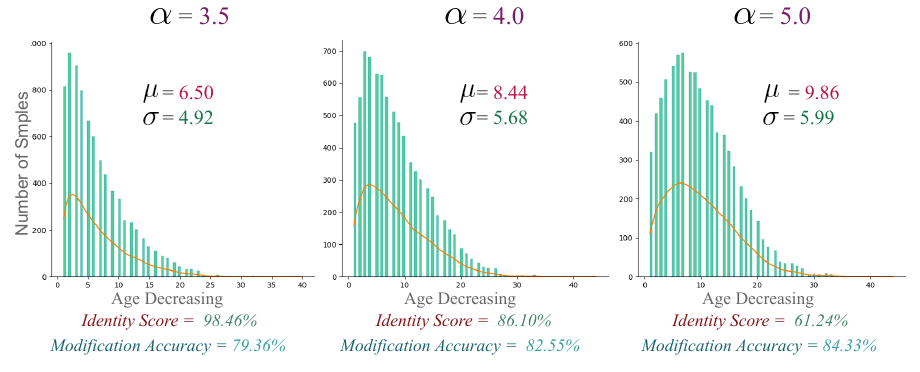}
  \caption{The evaluation of the transformation function for facial attribute editing with respect to \textit{Young} age. The figure shows the histogram of age differences, the modification accuracy, and the identity score for different values of $\alpha$.  The orange chart shows the Kernel Density Estimation of the histogram.}
  \label{fig:to_young_chart_identity}
\end{figure}

Next, we evaluated our transformation function for editing \textit{Man} attribute. As Table~\ref{tbl:identity_based_evaluation_fm} shows, setting $\alpha=2.5$, the identity score is $79.71\%$, while the facial attribute score is $94.39\%$. Increasing the value of $\alpha$ to $3.0$ reduces the identity score to $53.06\%$, while the facial attribute score increases to $99.36\%$. Finally, we increased the value of $\alpha$ to $4.0$, and around $99.60\%$ of the images in the validation set labeled as \textit{Man}, while the identity scores reduces to $41.89\%$. As expected, increasing the value of $\alpha$ increases the corresponding facial attribute score in the modified image, while the identity score reduces.

In another experiment, we evaluate our transformation function regarding the \textit{Anger} facial attribute modification. As Table~\ref{tbl:identity_based_evaluation_anger} shows, defining $\alpha=2.0$ results in the identity score of $98.87$, while the facial attribute modification score for \textit{Happy} to either \textit{Neutral} or \textit{Anger} is $61.64\%$ ($38.47\%$ for \textit{Happy} to \textit{Angry}, and $23.17\%$ for \textit{Happy} to \textit{Neutral}), and for \textit{Neutral} to \textit{Angry} is $69.50\%$. Increasing $\alpha$ to $2.5$ reduced the identity score to $84.67\%$, while increasing the facial attribute score for \textit{Happy} to \textit{Neutral/Anger} to $77.63\%$, and for \textit{Neutral} to \textit{Angry} to $78.00\%$. Following the same trend, setting $\alpha=3.0$ results in a reduction in identity score to $57.22\%$, and an increase in the facial attribute score: $87.70\%$ for \textit{Happy} to \textit{Neutral/Anger}, and $81.32\%$ for \textit{Neutral} to \textit{Angry}.

\begin{table}[b]
\caption{The evaluation of the transformation function for facial attribute editing with respect to \textit{Anger} expression.}
\label{tbl:identity_based_evaluation_anger}
\centering
\small
\resizebox{6cm}{!}{

\begin{tabular}{lccc}
\hline
                            & $\alpha=2.0$& $\alpha = 2.5 $       & $\alpha=3.0$        \\ \hline
Identity Score              & $98.87$         & $84.67$               & $57.22$     \\
Happy to Angry/Neutral(\%)  & $61.64$         & $77.63$               & $87.80$     \\
Happy to Angry (\%)         & $38.47$         & $65.21$               & $81.07$     \\
Happy to Neutral   (\%)     & $23.17$         & $14.59$               & $6.72 $    \\
Neutral to Angry  (\%)      & $69.50$         & $78.00$               & $81.32$     \\ \hline
\end{tabular}
}
\end{table}
\begin{table}[b]
\caption{The evaluation of the transformation function for facial attribute editing with respect to \textit{Black} race.}
\label{tbl:identity_based_evaluation_black}
\centering
\small
\resizebox{6cm}{!}{

\begin{tabular}{lccc}
\hline
                            & $\alpha= 1.5$     & $\alpha= 2.0$ & $\alpha=2.5$ \\ \hline
Identity Score (\%)         &  $84.67$      & $79.61$    &  $29.18$     \\
White to Black/Others (\%)   &  $35.41$      & $86.09$    &  $99.24$         \\
White to Black (\%)         &  $16.24$      & $76.56$    &  $98.42$          \\ 
White to Others (\%)         &  $19.16$      & $9.52 $    &  $0.81 $          \\ 
Others to Black (\%)         &  $43.27$      & $89.13$    &  $99.31$           \\ 
\hline
\end{tabular}
}
\end{table}

\begin{figure}[t!]
  \centering
  \includegraphics[width=\columnwidth]{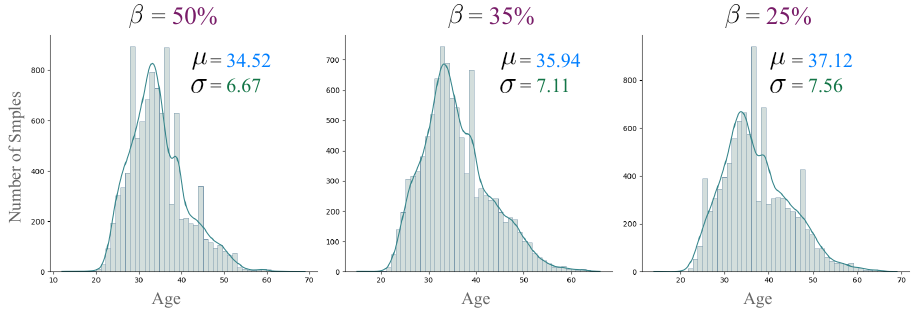}
  \caption{Age histogram of the validation set after using $\Psi$ for synthesis of the \textit{Old} faces. As the figure shows, as the amount of $\beta$ reduces, the mean of the generated images becomes older relative to the histogram of the original validation set before feature-base manipulation. The blue chart shows the Kernel Density Estimation of the histogram.}
  \label{fig:sem_to_old_hist}
\end{figure}

In the next experiment, we evaluated our method considering race modification to \textit{Black}. As Table~\ref{tbl:identity_based_evaluation_black} shows, for $\alpha=1.5$ the identity score is $84.67\%$, while the facial attribute score is around $35.41\%$ for \textit{White} to either \textit{Black} (around $16.24\%$) or \textit{Others} (around $19.16\%$) and $43.27\%$ for \textit{Others} to \textit{Black} race. As expected, increasing $\alpha$ to $2.0$, reduces the identity score to $79.61\%$, while facial attribute scores increase to $86.09\%$ and $89.13\%$ for \textit{White} to \textit{Black/Others}, and \textit{Others} to \textit{Black} respectively. Likewise, increasing $\alpha$ to $2.5$ results in a facial attribute score of $99.24\%$ for \textit{White} to \textit{Black/Others} race, out of which about $98.42\%$ of the samples with \textit{White} race converted to \textit{Black}. As mentioned in Sec~\ref{sec_tf_FAE}, increasing the value of $\alpha$ results in high-intensity facial attribute editing. We can consider \textit{White} to \textit{Black} race as high-intensity modification, which increased from $16.24\%$ for $\alpha=1.5$, to $76.56\%$ for $\alpha=2.0$, and finally, to $98.42\%$ for $\alpha=2.5$. On the contrary, \textit{White} to \textit{Others} race, which can be taken as low-intensity modification, reduces from $19.16\%$ for $\alpha=1.5$, to $9.52\%$ for $\alpha=2.0$, and finally, to $0.81\%$ for $\alpha=2.5$.

\begin{table}[b]
\caption{The evaluation of the transformation function for feature-based synthesis with respect to \textit{Man} as well as \textit{Woman} gender.}
\label{tbl:sem_based_evaluation_fm}
\centering
\small
\resizebox{6cm}{!}{

\begin{tabular}{lccc}
\hline
                         & $\beta= 25 \%$  & $\beta= 35 \%$ & $\beta=50 \%$ \\ \hline
Woman to Man (\%)      & $99.92$  & $99.86$      & $99.21$       \\ 
Man to Woman (\%)      & $98.88$  & $98.09$      & $96.60$       \\ \hline

\end{tabular}
}
\end{table}

\begin{table}[b]
\caption{The evaluation of the transformation function for feature-based synthesis with respect to \textit{Anger} expression. }
\label{tbl:sem_based_evaluation_anger}
\centering
\small
\resizebox{6cm}{!}{

\begin{tabular}{lccc}
\hline
                            & $\beta = 25\% $     & $\beta = 35\% $ & $\beta = 50\% $   \\ \hline
Happy to Angry/Neutral(\%)  &  $84.56$    & $79.55$      & $64.55$                     \\
Happy to Angry (\%)         &  $73.84$    & $62.33$      & $41.52$                     \\
Happy to Neutral   (\%)     &  $10.72$    & $17.21$      & $23.03$                     \\
Neutral to Angry  (\%)      &  $87.55$    & $83.81$      & $77.80$                     \\ \hline
\end{tabular}
}
\end{table}

\begin{table}[b]
\caption{The evaluation of the transformation function for feature-based synthesis with respect to \textit{Dark} skin color.}
\label{tbl:isem_based_evaluation_black}
\centering
\small
\resizebox{6cm}{!}{

\begin{tabular}{lccc}
\hline
                            & $\beta=25 \%$  & $\beta=35 \%$  & $\beta= 50\% $    \\ \hline
White to Black/Others (\%)   & $99.95$           & $99.97$          & $99.65$             \\
White to Black (\%)         & $99.93$           & $99.96$          & $99.32$             \\ 
White to Others (\%)         & $00.01$           & $0.01 $          & $0.32 $              \\ 
Others to Black (\%)         & $100  $           & $100  $          & $99.65$             \\ 
\hline
\end{tabular}
}
\end{table}

In order to evaluate our transformation function regarding age modification, we designed two sets of experiments. In the first set of experiments, we modified the latent vectors in the validation set to make the samples \textit{Older}. We defined the \textit{modification accuracy} by predicting the age of each sample, before and after the modification, and calculated the portion of samples where the age is greater after the modification. Fig.~\ref{fig:to_old_chart_identity} shows the histogram of the increase in age for different values of $\alpha$. For $\alpha=3.5$, the identity score is $96.30\%$ and the average increase in the age of the samples is around $8.46$ years. Increasing $\alpha$ to $4.0$ and $5.0$ results in the identity scores of $90.49\%$ and $63.83\%$ and the average increase in age by $9.71$ and $12.10$ years, respectively.

Similarly, we evaluated the transformation function for \textit{Young} age modification. As  Fig.~\ref{fig:to_young_chart_identity} shows, for $\alpha=3.5$, the identity score is $98.46\%$, the modification score $79.36\%$, and the average age reduction is around $6.50$ years. As expected, increasing $\alpha$ to $4.0$ and $5.0$ results in a reduction in identity scores to $86.10\%$ and $61.24\%$, an increased in the modification score to $82.55\%$ and $84.33\%$,
and the average decrease in age of $8.44$ and $9.86$ years, respectively. 

\begin{figure}[t]
  \centering
  \includegraphics[width=\columnwidth]{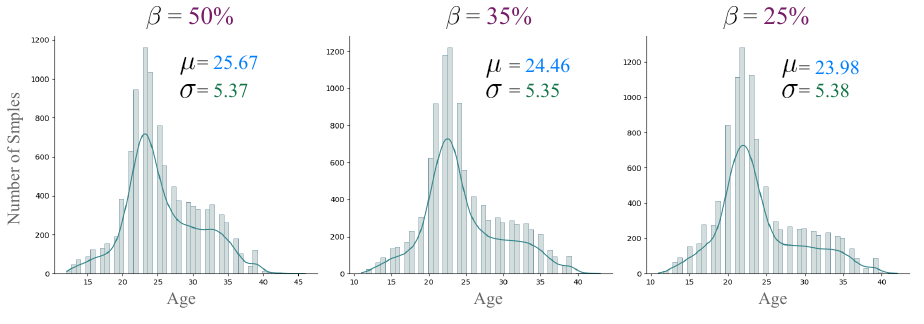}
  \caption{Age histogram of the validation set after using $\Psi$ for synthesis of the \textit{Young} faces. As the figure shows, as the amount of $\beta$ reduces, the mean of the generated images becomes younger relative to the histogram of the original validation set before feature-base manipulation. The orange chart shows the Kernel Density Estimation of the histogram.}
  \label{fig:sem_to_young_hist}
\end{figure}

\subsection{Feature-Based Synthesis}\label{eval_Semantic_Based_Synthesis}
To evaluate the transformation function for feature-based synthesis, we performed sets of experiments. In each experiment, we defined a target facial attribute $fa\_obj$ and manipulated a random latent vector such that its corresponding synthesized image inherits from $fa\_obj$. As Fig.~\ref{fig:evalset_histogram} shows, around $87\%$ of the randomly generated images using StyleGANs~\cite{karras2020training, karras2021alias} family are labeled as \textit{Happy} expression. For \textit{White} for ethnicity and \textit{Young} age, we have more than $83\%$, and $87\%$, respectively. Thus, we defined our target facial attributes as \textit{Neutral/Angry} for expression, \textit{Black} for ethnicity, \textit{Old} and \textit{Young} for age, and \textit{Woman} and \textit{Man} for gender. As explained in Sec.~\ref{sec_tf_SB}, for feature-based synthesis, we only measure the \textit{desired} facial attribute score of the synthesized image after the modification of the corresponding latent vector using our proposed transformation function, and preserving the original identity is not taken into account.

\begin{figure*}[t]
  \centering
  \includegraphics[width=18cm]{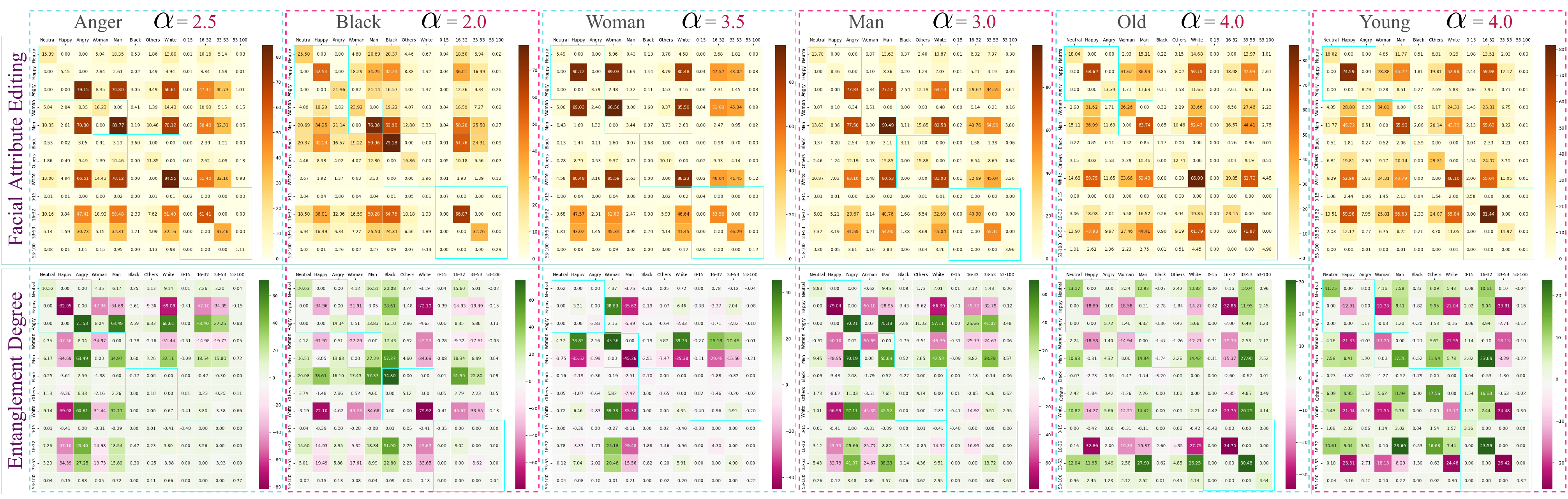}
  \caption{The first row shows the histogram of the validation set after facial attribute editing with respect to the mentioned facial attribute and the value of $\alpha$. In the second row, we show the Entanglement Degree, the difference between the original histogram (see Fig.~\ref{fig:heatmap_histogram}) of the validation set, and its corresponding histogram after we applied the transformation function. The results show the increase/decrease in the entanglement between different facial attributes.}
  \label{fig:hm_entanglement_degree}
\end{figure*}

In the first experiment, we manipulated the latent vectors in the validation set to generate images with \textit{Woman} gender. As Table~\ref{tbl:sem_based_evaluation_fm} shows, by decreasing the hyper-parameter $\beta$ from $50\%$ to $35\%$, and $25\%$, the corresponding facial attribute score increased from $96.60\%$ to $98.09\%$, and finally to $98.88\%$. Likewise, for \textit{Man} gender modification, for $\beta=50\%$, the facial attribute score is $99.21\%$. As expected, decreasing $\beta$ to $35\%$, and $25\%$ resulted in facial attribute score of $99.86\%$, $99.92\%$ respectively.

As Table~\ref{tbl:sem_based_evaluation_anger} shows, for the evaluation of \textit{Anger/Neutral} facial expression modification, we measure both \textit{Happy} to \textit{Angry}, as the high-intensity modification, and \textit{Happy} to \textit{Neutral}, as the low-intensity modification. As expected, reducing $\beta$ from $50\%$ to $35\%$, and $25\%$ results in increase of the facial attribute score from $64.55\%$ to $79.55\%$, and $84.56\%$ for \textit{Happy} to \textit{Anger/Neutral}, and $77.80\%$ to $83.81\%$, and $87.55\%$ for \textit{Neutral} to \textit{Angry} respectively. Furthermore, reducing the value of $\beta$, reduces the low-intensity modification (\textit{Happy} to \textit{Neutral}) score, while simultaneously increasing the high-intensity modification (\textit{Happy} to \textit{Angry}) score.

Table~\ref{tbl:isem_based_evaluation_black} shows the facial attribute score for race modification to \textit{Black}. As expected, reducing the value of $\beta$ increased the \textit{Black/Others} facial attribute score. Following the same trend, the facial attribute score for low-intensity modification (\textit{White} to \textit{Others}) decreased as $\beta$ decreased, and simultaneously we faced an increase in facial attribute score for high-intensity modification (\textit{White} to \textit{Black}).

Finally, we evaluated our transformation function for feature-based synthesis for age modification. For both \textit{Old} and \textit{Young} image synthesis, we followed our other experiments and set values of $\beta$ as $50\%$, $35\%$, and $25\%$. After the modification of the latent vectors in the validation set according to the desired \textit{facial attribute} (either \textit{Old} or \textit{Young}), we measure the age of the synthesized images and depict the corresponding histograms in Fig.~\ref{fig:sem_to_old_hist}, and Fig.~\ref{fig:sem_to_young_hist}. As Fig.~\ref{fig:sem_to_old_hist} shows, for $\beta$ sets to $50\%$, $35\%$, and $25\%$, the average age of the synthesized images is $34.52$ to $35.94$, and $37.12$ years old respectively, while the average age for the validation set is $31.53$ (see Fig.~\ref{fig:evalset_histogram}). For \textit{Young} facial attribute, as Fig.~\ref{fig:sem_to_young_hist}, decreasing $\beta$ from $50\%$ to $35\%$, and finally to $25\%$ results in a decrease in the average age of the synthesized images, from $25.67$ to $24.46$, and $23.98$ years old, respectively.

\subsection{Disentanglement Analysis and Attribute Editing} \label{sec_eval_DisentanglementAnalysis}
As we mentioned in Sec.~\ref{sec_Semantic_Entanglement_and_Solution}, facial attribute entanglement in the latent space of GANs would negatively affect facial attribute editing. In this section, we introduce experiments to evaluate our proposed solution for \textit{disentangled} facial attribute editing.

To show facial attribute entanglement in facial attribute editing, in Fig.~\ref{fig:hm_entanglement_degree}, we depict the Covariance matrix of the validation set regarding different facial attributes, for each of the following facial attributes: \textit{Anger} with $\alpha=2.5$, \textit{Black} race with $\alpha=2.0$, \textit{Woman} with $\alpha=3.5$, \textit{Man} with $\alpha=3.0$, \textit{Old} with $\alpha=3.5$, and \textit{Young} with $\alpha=4.0$. The second row of Fig.~\ref{fig:hm_entanglement_degree} shows the \textit{Entanglement Degree} figure, which is the difference between the Covariance matrix of the validation set before any modification and the Covariance matrix of the validation set after facial attribute editing with respect to the mentioned facial attributes. Accordingly, the positive values show the \textit{direct} entanglement, while the negative values show the \textit{reverse} entanglement. Based on Fig.~\ref{fig:hm_entanglement_degree}, we can express the following entanglements: 1- \textit{Anger} has direct entanglement with the \textit{Man}, and inverse entanglement with \textit{Woman}. 2- \textit{Black} race has direct entanglement with the \textit{Man} and \textit{Anger}, and inverse entanglement with \textit{Woman} and \textit{Happy}. 3-  \textit{Man} has direct entanglement with the \textit{Anger}, and inverse entanglement with \textit{Happy}. 4-  3-  \textit{Old} has direct entanglement with the \textit{Anger/Neutral} and \textit{Man}, and inverse entanglement with \textit{Happy} and \textit{Woman}. 5- \textit{Young} has direct entanglement with the \textit{Neutral} and \textit{Man}, and inverse entanglement with \textit{Happy} and \textit{Woman}. 6- On the contrary, as we also showed in Fig.~\ref{fig:mean_entanglement}, \textit{Woman} facial attribute has almost no entanglement with the other facial attributes. 

\begin{figure}[t]
  \centering
  \includegraphics[width=\columnwidth]{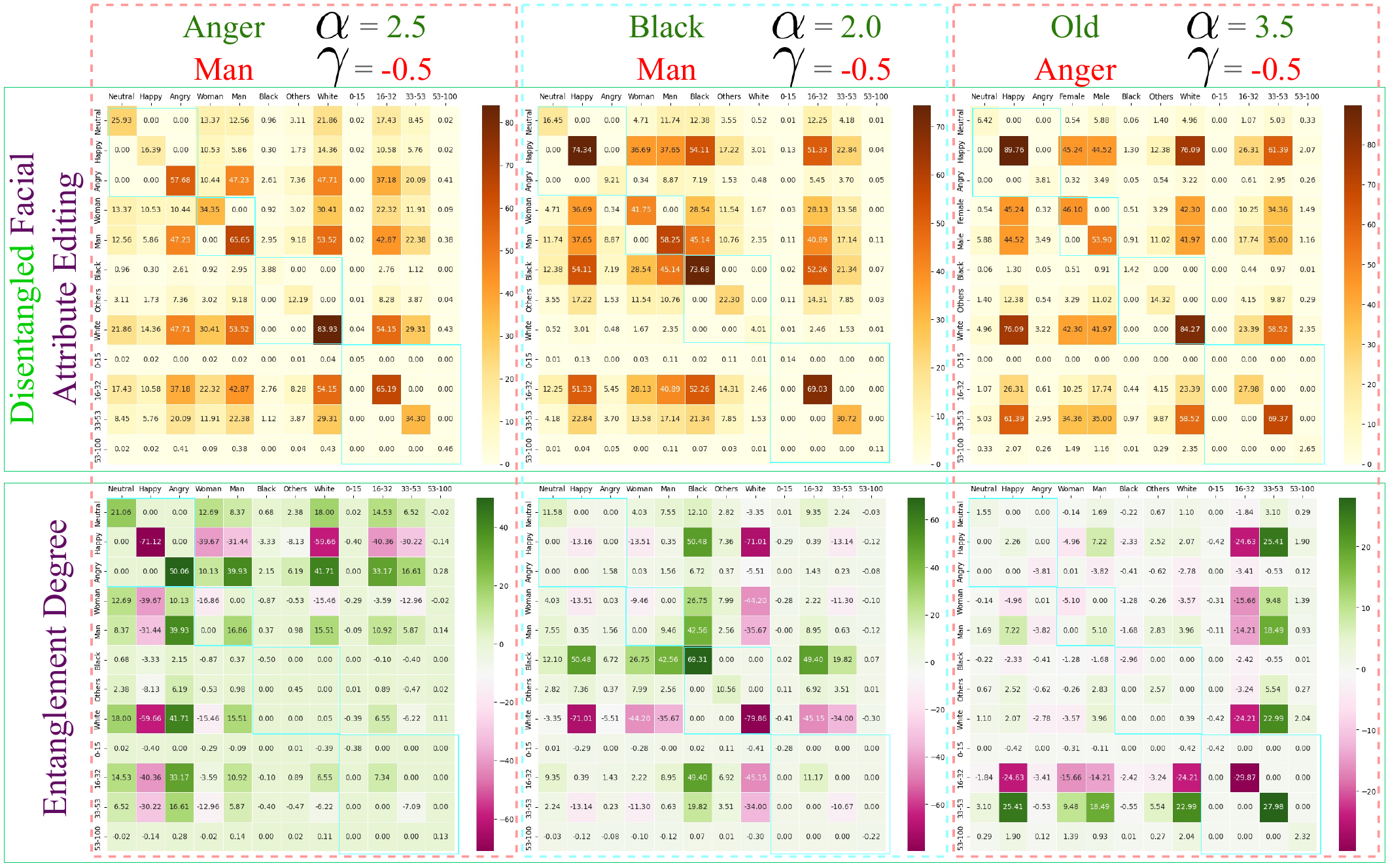}
  \caption{The first row shows the histogram of the validation set after the \textit{disentangled} facial attribute editing with respect to the mentioned facial attribute and the value of $\alpha$, and $\gamma$. In the second row, we show the Entanglement Degree, the difference between the original histogram (see Fig.~\ref{fig:heatmap_histogram}) of the validation set, and its corresponding histogram after we applied the disentangled transformation function.}
  \label{fig:hm_disentanglement_degree}
\end{figure}

In order to evaluate our proposed solution for disentangled facial attribute editing (see Sec.~\ref{sec_Semantic_Entanglement_and_Solution}), we perform 3 experiments for modification of \textit{Anger} (entangled with \textit{Man}), \textit{Black} (entangled with \textit{Man}), and \textit{Man} (entangled with \textit{Anger}). We depict the Covariance matrix of the validation set after modification and the corresponding \textit{Entanglement Degree} figure.

In the first experiment, we define our \textit{desired} facial attribute, $fa\_obj_D$, as \textit{Anger} facial attribute with $\alpha=2.5$, and set the entangled facial attribute, $fa\_obj_U$, as \textit{Man} with $\gamma= -0.5$ (see Eq.~\ref{eq:transformation_function_disentangled}). As Fig.~\ref{fig:hm_disentanglement_degree} shows, the entanglement degree between \textit{Anger}, and \textit{Man} facial attributes reduced dramatically. 

In the second experiment, we define our \textit{desired} facial attribute, $fa\_obj_D$, as \textit{Black} with $\alpha=2.0$, and set the entangled facial attribute, $fa\_obj_U$, as \textit{Man} with $\gamma= -0.5$. As Fig.~\ref{fig:hm_disentanglement_degree} shows, after the disentangled synthesis of the images, there is almost no entanglement between \textit{Black}, and \textit{Man} facial attributes.

In the third experiment, we define our \textit{desired} facial attribute, $fa\_obj_D$, as \textit{Man} facial attribute with $\alpha=3.0$, and set the entangled facial attribute, $fa\_obj_U$, as \textit{Anger} with $\gamma= -0.5$. As Fig.~\ref{fig:hm_disentanglement_degree} shows, after the disentangled synthesis of the images, there is almost no entanglement between \textit{Man}, and \textit{Anger} facial attributes. However, as expected, \textit{Man} facial attribute still is entangled with the age, and we observe an increase in the age of the synthesized images.

\begin{figure}[t]
  \centering
  \includegraphics[width=\columnwidth]{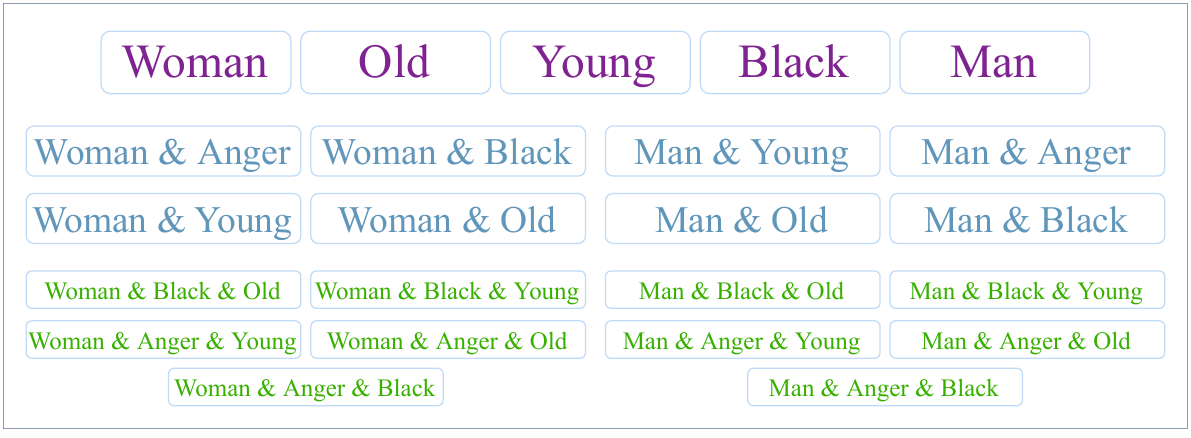}
  \caption{ We create our dataset using 23 different combinations of the least existing facial attribute. For each combination, we generate 2K images, which results in 46K images in total.  }
  \label{fig:dataset_config}
\end{figure}

\subsection{Generated Dataset} \label{sec_proposed_dataset}
Our proposed transformation function for multiple feature-based syntheses is a powerful tool to define a diverse dataset considering human expression, age, gender, and race. To create the dataset, we first defined a set of desired facial attributes (\textit{e.g} {woman, Black, Angry}), and then create the corresponding transformation function for multiple feature-based syntheses (see Sec.~\ref{sec_multiple_Semantic_Manipulation}). For each set, we randomly sample 2K latent vectors, manipulate them using the corresponding transformation function, and finally synthesize the corresponding images using the generative function $\mathbf{G}$.

\begin{figure}[t]
  \centering
  \includegraphics[width=\columnwidth]{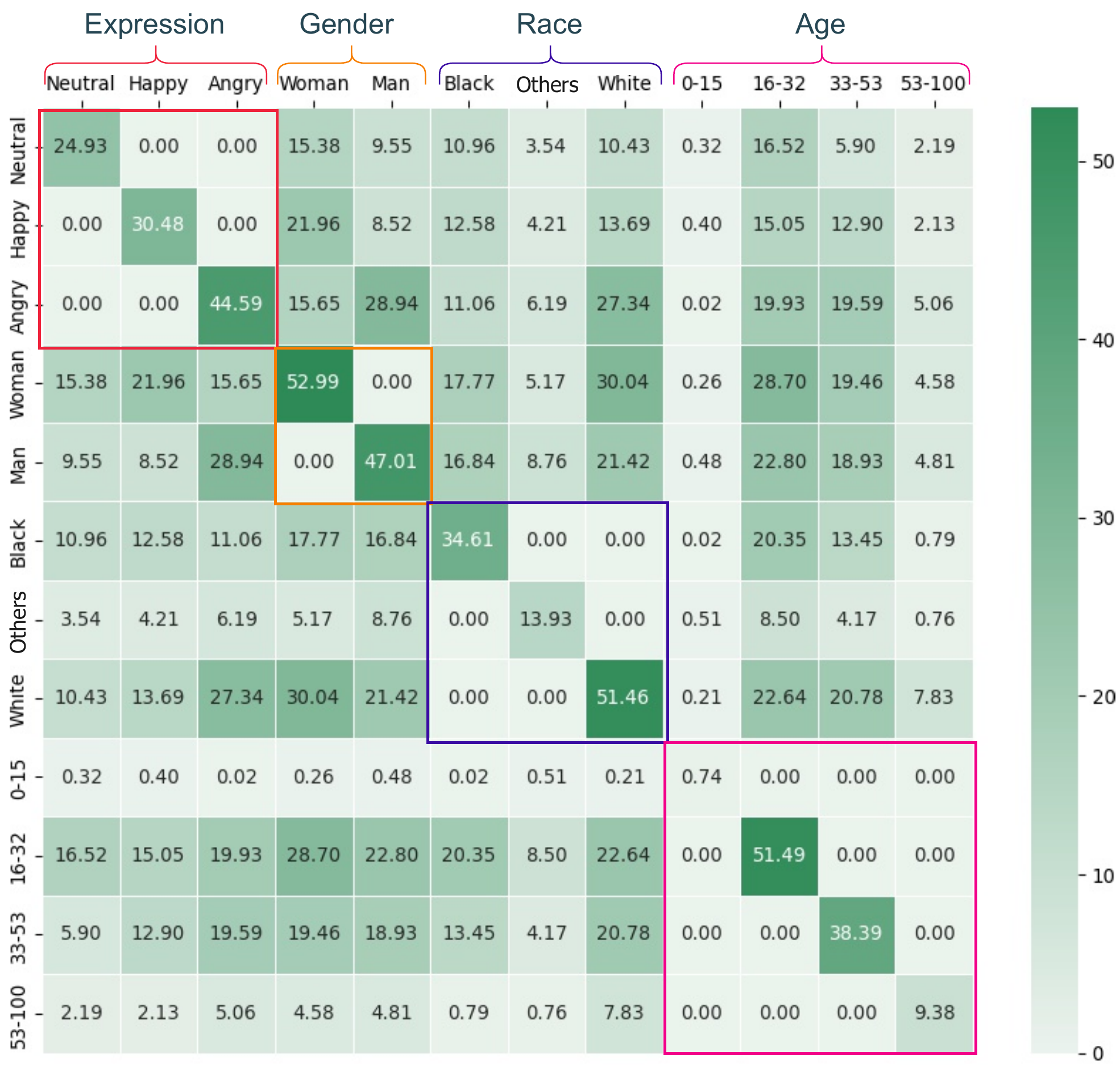}
  \caption{The Covariance Matrix of our proposed balanced dataset. The red box shows the histogram of the facial emotion expression: $24.93\%$ of all samples are Neutral, $30.48\%$ Happy, and $44.59\%$ Angry. The orange box shows the histogram of the gender: $52.99\%$ of samples are women, and $47\%$ are men. The blue box shows the histogram of the race: $34.61\%$ of samples are Black, $13.93\%$ Others ($48.6\%$ are Black in total), and $51.46\%$ are White. The pink box shows the histogram of the age: around $53\%$ of the sample is less than 32 years old, while the rest are older than 32.}
  \label{fig:merged_covm}
\end{figure}

In Fig.~\ref{fig:dataset_config}, we proposed each of our defined sets of desired facial attributes. While we have not introduced each possible combination of facial attributes, we tried our best to define each subset such that the final proposed dataset becomes as diverse as possible. As Fig.~\ref{fig:dataset_config} shows, we have defined 23 different combinations, and hence, the final dataset includes 46K images, their corresponding latent vector, and annotations. In order to evaluate the diversity and the balance of the dataset, we depict its 
Covariance Matrix of its facial attributes in Fig.~\ref{fig:merged_covm}. Compared to our validation set (see Fig.~\ref{fig:heatmap_histogram}) where about $87.51\%$ of the samples are Happy, the number of Happy samples reduced to about $30.48\%$. For the Neutral, and Angry expressions, we only have $4.87\%$, and $7.62\%$ of the sample in the validation set, while these ratios increased dramatically to $24.93\%$, and $44.59\%$ in our proposed dataset. Considering gender, Fig.~\ref{fig:merged_covm} shows that we have almost a balanced combination between women ($52.99\%$), and man gender($47.01\%$). Likewise, considering race, $83.88\%$ of samples in the validation set are White, while in our proposed dataset this ratio reduced dramatically to $51.46\%$. Similarly, we only have $4.38\%$ and $11.75\%$ Black and Others in the validation set respectively, while these ratios increased dramatically to $34.61\%$, and $13.893\%$ in our proposed dataset. For the age, we witness that the Old samples dramatically increased from less than $1\%$ in the validation set to around $9.38\%$ in our proposed dataset.

Another important attribute of our proposed dataset is its diversity within multiple facial attribute classes (\textit{e.g. Old Anger woman}). By looking at Fig.~\ref{fig:heatmap_histogram}, the Covariance Matrix of facial attributes in the validation set, and comparing it to Fig.~\ref{fig:merged_covm}, we can see that the former is much more sparse than the latter. The sparsity in the Covariance Matrix of facial attributes explicitly shows how imbalanced a dataset is. To illustrate, as Fig.~\ref{fig:heatmap_histogram} shows, $87.51\%$ of samples in the validation set are Happy, $74.02\%$ of samples are Happy-White, and only $3.63\%$ and $9.86\%$ of samples are Happy-Black, and Happy-Others respectively. In contrast, in our proposed dataset, around $30.48\%$ of samples are Happy, only $13.69\%$ of samples are Happy-White, and we have $12.58\%$ and $4.21\%$ of sample Happy-Black, and Happy-Others respectively. In other words, while in the validation set (which is generated using StyleGANs~\cite{karras2020training, karras2021alias} family with no modification in the latent space) we can rarely find a Happy sample with Black, in our proposed dataset, more than half of the Happy samples are Black. Another good example is the woman, where almost all woman samples in the validation set are Happy (about $51.20\%$ are women in total, and $50.20\%$ are woman-Happy, and less than $1\%$ woman-Angry and woman-Neutral), in our proposed dataset, about $52.99\%$ of samples are woman, we have $21.15\%$ woman-Happy, and $15.65\%$ and $15.38\%$ woman-Angry and woman-Neutral respectively. Exploring the Covariance Matrix of the facial attributes of our proposed dataset in Fig.~\ref{fig:merged_covm}, and comparing it to that of the validation set depicted in Fig.~\ref{fig:heatmap_histogram}, supports our claim about how diverse and balanced our proposed dataset is. We have provided detailed information about each subset of our proposed dataset in supplementary materials.

\section{Discussion and Conclusion}\label{sec_conclusion}
Since GANs usually do not provide any information about the relation between the input vector, and the facial attributes and attributes of the synthesized image, interpreting the latent space of GANs is a vital step in controlling the features of the image generation. There are two main approaches for analyzing and manipulating the latent space of a well-trained GAN: \textit{supervised and unsupervised analyses}. Our approach belongs to the supervised analysis category. The main drawback of the supervised methods, where off-the-shelf classifiers are used to label the synthesized images, is that they heavily rely on the performance of the utilized classifiers. However, compared to the unsupervised method, where the similarity between the latent vectors is captured and used for latent space analysis, the supervised approaches provide more specific control over the latent space manipulation, since the desired target attributes can be defined precisely. A combination of these approaches is needed and worth further exploration.

Besides, per category feature imbalance of existing datasets that are mostly being used for training facial GANs, impacts the GANs drastically and causes attribute entanglement. As discussed and shown in Sec.~\ref{sec_evaluation}, not only facial GANs can hardly synthesize some specific combination of facial attributes (\textit{e.g.} an old black woman) without latent code modification, but also modification of specific facial attributes mostly affects the other entangled feature of a face too (\textit{e.g.} the entanglement between Anger and man attributes). Training or fine-tuning GANs on a balanced dataset, or penalizing GANs during the training process to generate a more diverse combination of facial attributes should be taken into account to deal with the mentioned drawbacks of facial GANs.

In conclusion, we presented a framework for analyzing and manipulating the latent space of well-trained GANs. First, we randomly synthesized $100K$ images using StyleGANs~\cite{karras2020training, karras2021alias} family as our training set. For any image in the training set, we utilized 4 different off-the-shelf classifiers to predict the facial expression, age, gender, and race. Then, using the Eigenvectors of the Covariance matrix of the latent vectors having a specific facial attribute (\textit{e.x.} Anger), we proposed a transformation function for both single and multiple facial attribute editing and feature-based synthesis. 
 
We also showed that due to the implicit entanglement in the training set of GANs, they usually entangle specific facial attributes and features with each other. Hence, we analyzed facial attribute entanglement in the latent space of GANs and provided an effective solution for highly disentangled facial attribute editing. Our evaluations show that our proposed GANalyzer framework can be utilized for accurate facial attribute editing, and feature-based synthesis in a wide range of applications. Finally, by utilizing our proposed framework, we generated a diverse photo-realistic human facial dataset with 23 different combinations of facial attributes. Our generated dataset which contains 46K images and the corresponding annotations, would be beneficial to many other automatic recognition tasks as well as physiological studies. Despite it is showed that our proposed dataset is relatively diverse, due to the fact that the image annotation has been done using deep learning-based algorithms, it is worth using human annotators for more accurate, and trustable annotation of the dataset.

\section*{Acknowledgment}
This work is partially supported by NSF grants (2141313 and 2141289), and an internal PROF grant from the University of Denver.

\ifCLASSOPTIONcaptionsoff
  \newpage
\fi


\bibliographystyle{IEEEtran}
\bibliography{references}

\vspace{-2.5cm}

\begin{IEEEbiography}[{\includegraphics[width=1in,height=1.25in,clip,keepaspectratio]{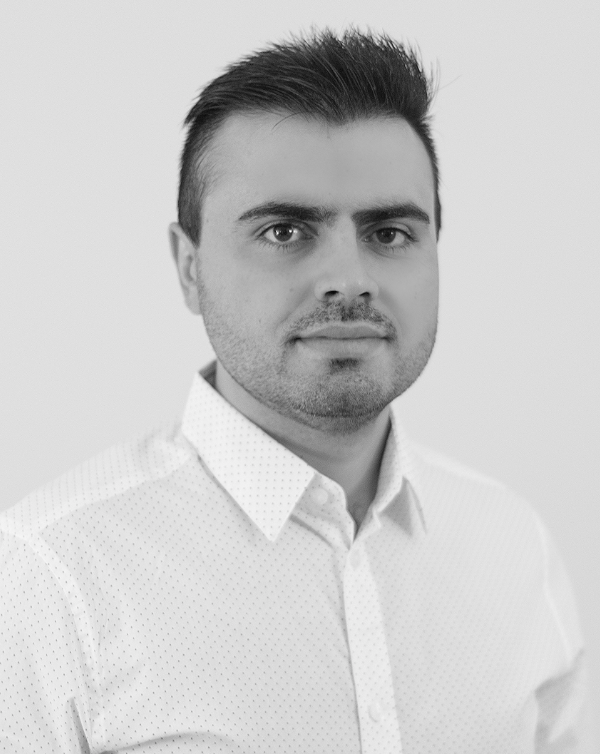}}]{Ali Pourramezan Fard} received the MSc degree in Computer Engineering - from Iran University of Science and Technology, Tehran, Iran, in 2015. He is currently pursuing his Ph.D. degree in Electrical \& Computer engineering and is a graduate research assistant in the Department of Electrical and Computer Engineering at the University of Denver. His research interests include Computer Vision, Machine Learning, and Deep Neural Networks, especially in face alignment, and facial expression analysis.
\end{IEEEbiography}

\vspace{-1.9cm}

\begin{IEEEbiography}[{\includegraphics[width=1in,height=1.25in,clip,keepaspectratio]{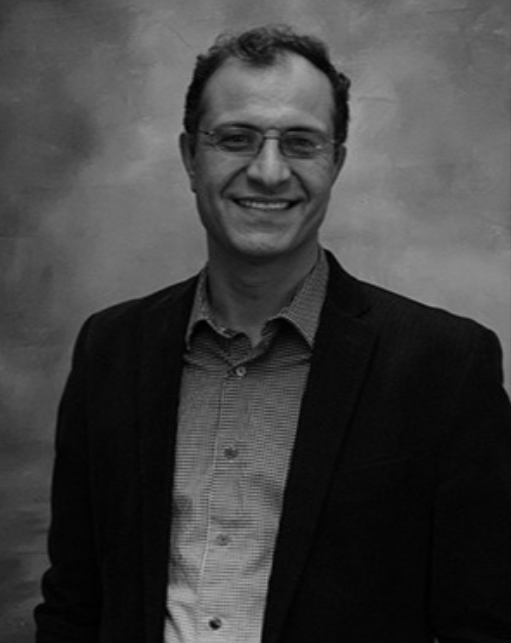}}]{Mohammad H. Mahoor} received the MS degree in Biomedical Engineering from Sharif University of Technology, Iran, in 1998, and the Ph.D. degree in Electrical and Computer Engineering from the University of Miami, Florida, in 2007. Currently, he is a professor of Electrical and Computer Engineering at the University of Denver. He does research in the area of computer vision and machine learning including visual object recognition, object tracking, affective computing, and human-robot interaction (HRI) such as humanoid social robots for interaction and intervention of children with autism and older adults with depression and dementia. He has received over \$7M in research funding from state and federal agencies
including the National Science Foundation and the National Institute of Health. He is a Senior Member of IEEE and has published over 158 conference and journal papers.
\end{IEEEbiography}

\vspace{-1.9cm}

\begin{IEEEbiography}[{\includegraphics[width=1in,height=1.25in,clip,keepaspectratio]{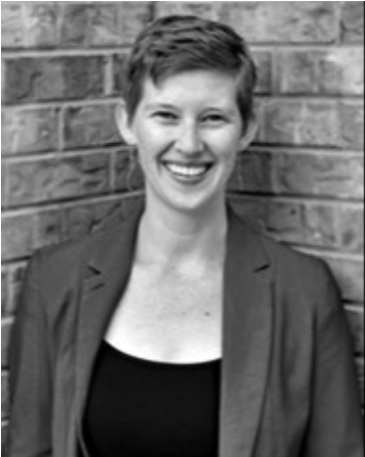}}]{Sarah Ariel Lamer} earned her Ph.D. in Social Psychology from the University of Denver in 2019. She is now an Assistant Professor of Social Psychology at the University of Tennessee in Knoxville where she studies how adults and children learn stereotypes from patterns that are present in their culturally shared environments. She and her lab draw from a variety of methods to explore this topic including psychophysics, drift-diffusion modeling, and representative sampling. Her lab is supported by funding from the National Science Foundation and the Research for Indigenous Social Action and Equity Center. 
\end{IEEEbiography}

\vspace{-13.9cm}

\begin{IEEEbiography}[{\includegraphics[width=1in,height=1.25in,clip,keepaspectratio]{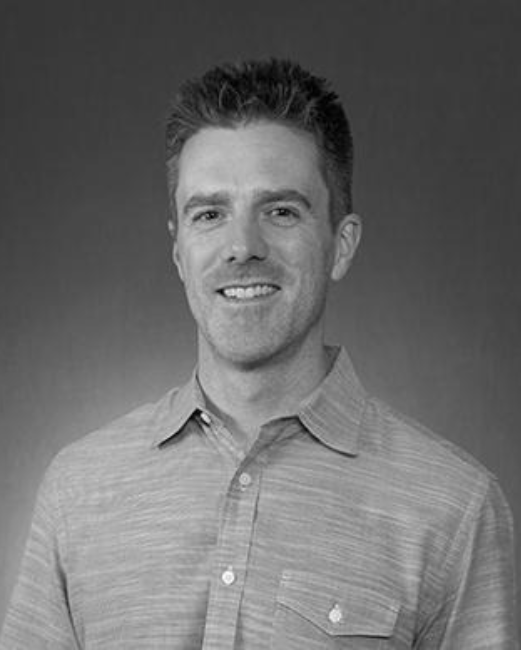}}]{Timothy Sweeny} received a Ph.D. in Psychology from Northwestern University in Evanston, Illinois, in 2010. He received postdoctoral training in the Department of Psychology at the University of California, Berkeley, from 2010-2013. Currently, he is an Associate Professor in the Department of Psychology at the University of Denver. He conducts research at the intersection of vision science and social psychology, with an emphasis on visual awareness, organization, as well as the perception of emotion, crowds, and gaze. He conducts research with support from the National Science Foundation and the National Institute of Health.
\end{IEEEbiography}

\end{document}